%% file: main.tex
\PassOptionsToPackage{table,dvipsnames}{xcolor}
\newif\ifappleformat
\appleformattrue % set to \appleformatfalse for Interspeech
\ifappleformat
  \documentclass{templates/apple/applemlr}
\else
  \documentclass[cameraready]{templates/interspeech2026/Interspeech}
\fi

\input{preamble}

\input{paper_metadata}
\input{title_authors}
\input{abstract}
\ifappleformat
  \setcitestyle{numbers,square}
\fi

\begin{document}
\maketitle
\ifappleformat\else
\begin{abstract}
\paperabstract
\end{abstract}
\fi
\input{body/03_01_introduction}
\input{body/03_02_related_work}
\input{body/03_03_experiments}
\FloatBarrier

\section{Conclusion}
We present the first scaling law analysis for continuous diffusion spoken language models trained without text supervision. Validation loss and our proposed pJSD metric for ``languageness'' both follow power-law behavior, mirroring AR SLM trends. 
The optimal token-to-parameter ratio decreases with compute, indicating improved data efficiency at scale.
Moreover, higher computes allow near-optimal performance across a much wider variety of parameter-to-data allocations, opening up possibility for fast inference.
% , though a gap relative to AR models persists. 
Most perceptual quality metrics saturate near real-data baselines and lack scaling laws; 
for the Meta Audiobox Aesthetics components that our fused two-stage scaling law fits suggests baseline performance may remain unreachable through scaling alone. 
Ablations show data scale drives linguistic quality while noise schedule governs perceptual fidelity. Scaling to 16B parameters produces emotive, multi-speaker, multilingual speech, yet long-form coherence remains elusive, suggesting that closing the gap with text-based models requires advances in speech representations or joint text-speech modeling rather than further scaling.
\section{Generative AI Use Disclosure}
Generative AI tools were used for language editing and polishing the manuscript text. 
AI tools were also used to clean-up, document, and type-annotate hand-written scripts used in generation of results that are added to the paper.
Scientific content, experimental design and results, analysis, implementation, or conclusions are the sole work of the authors. 
All authors take full responsibility for the content of the manuscript.
% % [Eeshan]: I think this is worth mentioning?
% No generative AI was used to produce experimental results, figures, or tables. 

% Bib

% [Eeshan]: Pierre Ablin's original scaling law fit was given to me by Jason; I have modified it heavily but would like to acknowledge because I kept the core of that script for scaling law fitting.
% [Eeshan]: Louis B\'{e}thune's script `style.py` from `primus_plots`, which didn't work off the shelf, but was really useful upon grokking to style our plots. 
\section{Acknowledgment}
We thank Pierre Ablin, Zak Aldeneh, Richard He Bai, Samy Bengio, Kari Nouriy, Timea Kutasi, Barry Theobald, Ruixiang Zhang for helpful discussions; Vivek Kumar, Sanskruti Shah, Shaoen Qin for help with data.
Names are in alphabetical order by last name within the group.

\ifappleformat
  \setlength{\bibsep}{0pt}
  \renewcommand{\bibfont}{\small}
  \bibliography{main}
  \bibliographystyle{unsrtnat}
\else
  \clearpage
  \makeatletter
  \def\eightpt{\fontsize{7.7}{8.05}\selectfont}
  \makeatother
  \bibliography{main}
  \bibliographystyle{templates/interspeech2026/IEEEtran}
\fi

% \clearpage

\section{Contributions}

\begin{itemize}
    \item \textbf{Implementation} All code for model training is written by Jason Ramapuram. All code for model evaluation (including metrics implementation) is written by Eeshan Gunesh Dhekane in consultation with Russ Webb, Jason Ramapuram, Tatiana Likhomanenko, Navdeep Jaitly.
    \item \textbf{Data}~~All data are prepared by Tatiana Likhomanenko with help on data filtering and debugging from Zijin Gu.
    \item \textbf{Model Design}~~Model design is done by Jason Ramapuram in consultation with Navdeep Jaitly and Tatiana Likhomanenko.
    \item \textbf{Training}~~All models for scaling law fits are trained by Eeshan Gunesh Dhekane in consultation with Jason Ramapuram.
    \item \textbf{Evaluation}~~All model evaluation is executed by Eeshan Gunesh Dhekane.
    \item \textbf{Scaling Law}~~Scaling law analysis and fits are instrumented by Eeshan Gunesh Dhekane in consultation with Amitis Shidani, Tatiana Likhomanenko, Jason Ramapuram and Russ Webb.
    \item \textbf{Phoneme Jensen-Shannon Divergence (pJSD) Metric}~~pJSD is proposed during discussion between Navdeep Jaitly, Dan Busbridge, Tatiana Likhomanenko and Jason Ramapuram. 
    \item \textbf{Ablations}~~All ablation models are designed and trained by Jason Ramapuram.
    \item \textbf{Scaling to 16B models}~~Scaling to 16B model including the model design is done by Jason Ramapuram.
    \item \textbf{Classifier Guidance for ASR}~~Investigations and code looking into classifier guidance using an ASR model as guidance were conducted by Zijin Gu, but was dropped for the final manuscript.
    \item \textbf{RLAIF}~~Investigations and code for RL with AI feedback were written by Bogdan Mazoure, but was dropped for final manuscript.
    \item \textbf{Writing and Paper Preparation}~~The manuscript was jointly written by Tatiana Likhomanenko, Jason Ramapuram, Amitis Shidani, and Eeshan Gunesh Dhekane. It was edited and reviewed by all other authors.
    \item \textbf{Advising} Navdeep Jaitly and Tatiana Likhomanenko advised other co-authors at every stage of the project. Navdeep Jaitly was the original project driver, while Tatiana Likhomanenko took over at later stages to help aggregate and execute results into the publication and frame the work.
\end{itemize}

% [Resolved]: Results and discussions based on points of the quip.
%   The results whose gist is added into main body are commented.
%   These serve as ideal starting point for an arxiv later on.
% \input{body/98_temporary_devlog_isoFLOPs_analysis}
% \input{body/97_temporary_devlog_fitting_scaling_laws}

% NB(jrp): cutting for submission. Bring back for arXiv
% \input{body/99_appendix}
\end{document}

%% file: preamble.tex
\ifappleformat
  \input{preamble/apple_preamble}

\else
  \input{preamble/interspeech_preamble}
\fi

\input{preamble/packages}

\input{preamble/style}

\input{preamble/symbols}

\input{preamble/math}

%% file: preamble/apple_preamble.tex
\usepackage{amsmath}
\usepackage{enumerate}
\usepackage{algorithm}
\usepackage{algpseudocode}
\usepackage{amsfonts}
\usepackage{amsthm}
\usepackage{newtxtt}
\usepackage{cleveref}
\usepackage{diagbox}
\usepackage{colortbl}
\usepackage{amssymb}
\usepackage{xspace}
\usepackage{wrapfig}
\usepackage{adjustbox}
\usepackage{tabularx}
\usepackage{booktabs}
\usepackage{mathtools}
\usepackage{tikz}
\usepackage{silence}
\usepackage{dsfont}
\usepackage[table]{xcolor}
\usepackage[dvipsnames]{xcolor}
\usepackage{multirow}
\usepackage{makecell}
\usepackage{xfakebold}

\definecolor{textgray}{HTML}{6E6E73}
\usetikzlibrary{positioning, calc}
\usetikzlibrary{decorations.pathmorphing}

\makeatletter
\patchcmd{\wrong@fontshape}{\@gobbletwo}{}{}{}
\makeatother
\numberwithin{equation}{section}
\makeatletter
\AtBeginDocument{
  \urlstyle{sf}
  
}
\makeatother

\definecolor{light}{RGB}{125, 125, 125}
\crefname{tcb@cnt@pbox}{code}{code}
\Crefname{tcb@cnt@pbox}{Code}{Code}
\crefname{assumption}{assumption}{assumption}
\Crefname{assumption}{Assumption}{Assumptions}

\newtcolorbox[auto counter]{pbox}[2][]{
  colback=white,
  title=Code~\thetcbcounter: #2,
  #1,fonttitle=\sffamily,
  fontupper=\sffamily,
  arc=2pt,
  colframe=bgcolor,
  coltitle=fgcolor,
  colbacktitle=bgcolor,
  toptitle=0.25cm,
  bottomtitle=0.125cm
}

\makeatletter
\newcommand\applefootnote[1]{%
  \begingroup
  \renewcommand\thefootnote{}%
  \renewcommand\@makefntext[1]{\noindent##1}%
  \footnote{#1}%
  \addtocounter{footnote}{-1}%
  \endgroup
}
\makeatother

\definecolor{cverbbg}{gray}{0.90}

%% file: preamble/interspeech_preamble.tex
\usepackage{float}
\usepackage{placeins}

%% file: preamble/packages.tex
\usepackage{amsmath, amsthm, amssymb}

\usepackage{expl3}  % https://tex.stackexchange.com/questions/20456/removing-a-backslash-when-passing-arguments-to-newcommand
\ExplSyntaxOn
  _to_str:N
\ExplSyntaxOff
\usepackage{ifthen}
\usepackage{pgffor}

\usepackage[acronym,smallcaps,nowarn,section,nonumberlist]{glossaries}[=v4.46]
\glsdisablehyper  % We don't want usually want hyperlinks in  papers as there's no glossary to link to.
\makeglossaries

\setacronymstyle{long-short}

\loadglsentries{preamble/acronyms.tex}

\usepackage{cleveref}
\usepackage{bm}
\usepackage{float}
\usepackage{placeins}
\ifappleformat
  \usepackage[T1]{fontenc}  % For some non-latin characters.
  \usepackage[type1]{libertine}
  \renewcommand{\sfdefault}{sfpro}
  \usepackage[libertine]{newtxmath}
\fi
\usepackage{mathtools}

\usepackage{booktabs} 
\usepackage{multirow}
\usepackage{subcaption}

%% file: preamble/acronyms.tex
% Some commonly used acronnyms.
% Please add any new ones here.
% Try to maintain (approximate) alphabetical ordering :).

% A
\newacronym{auc}{AUC}{Area Under the Curve}

% B
\newacronym{bert}{BERT}{Bidirectional Encoder Representations from Transformers}
\newacronym{bow}{BOW}{Bag of Words}
\newacronym{boe}{BOE}{Bag of Embeddings}
\newacronym{borep}{BOREP}{Bag of Random Embedding Projections}
\newacronym{bn}{BN}{Bayesian Network}

% C
\newacronym{cdf}{CDF}{Cumulative Distribution Function}
\newacronym{cnn}{CNN}{Convolutional Neural Network}
\newacronym{cka}{CKA}{Centered Kernel Alignment}
\newacronym{cvae}{CVAE}{Conditional Variational Auto-Encoder}
\newacronym{cfg}{CFG}{Classifier free guidance}

% D
\newacronym{dag}{DAG}{Directed Acyclic Graph}
\newacronym{dgm}{DGM}{Directed Graphical Model}
\newacronym{dnn}{DNN}{Deep Neural Network}
\newacronym{dit}{DiT}{diffusion transformers}

% E
\newacronym{ecfp4}{ECFP4}{Extended Connectivity Fingerprint}
\newacronym{ehr}{EHR}{Electronic Health Record}
\newacronym{esn}{ESN}{Echo State Network}

% F
\newacronym{fft}{FFT}{Fast Fourier Transform}
\newacronym{fs}{FS}{FastSent}
\newacronym{ft}{FT}{Fourier Transform}

% G
\newacronym{gat}{GAT}{Graph Attention Network}
\newacronym{gcn}{GCN}{Graph Convolutional Network}
\newacronym{gdl}{GDL}{Geometric Deep Learning}
\newacronym{ggnn}{GGNN}{Gated Graph Neural Network}
\newacronym{glove}{GloVe}{Global Vectors for Word Representation}
\newacronym{gnn}{GNN}{Graph Neural Network}
\newacronym{gru}{GRU}{Gated Recurrent Unit}

% H
\newacronym{hmm}{HMM}{Hidden Markov Model}

% I
\newacronym{idf}{IDF}{Inverse Document Frequency}
\newacronym{ig}{IG}{Information Gain}
\newacronym{isan}{ISAN}{Input Switched Affine Network}

% J
\newacronym{jsd}{JSD}{Jensen-Shannon Divergence}

% K
\newacronym{kld}{KLD}{Kullback-Leibler Divergence}
\newacronym{knn}{KNN}{$K$-Nearest Neighbour}

% L
\newacronym{lhs}{L.H.S.}{left hand side}
\newacronym{lstm}{LSTM}{Long Short Term Memory Network}

% M
\newacronym{map}{MAP}{Maximum a Posteriori}
\newacronym{mc}{MC}{Monte Carlo}
\newacronym{mdl}{MDL}{Minimum Description Length}
\newacronym{ml}{ML}{Machine Learning}
\newacronym{mlp}{MLP}{multi-layer perceptron}
\newacronym{mle}{MLE}{Maximum Likelihood Estimation}
\newacronym{mnist}{MNIST}{Modified National Institute of Standards and Technology}
\newacronym{mmdit}{MM-DiT}{multimodal diffusion transformer}

% N
\newacronym{nlp}{NLP}{Natural Language Processing}
\newacronym{nn}{NN}{Neural Network}
\newacronym{nll}{NLL}{Negative Log Likelihood}

% O
\newacronym{ode}{ODE}{Ordinary Differential Equation}
\newacronym{oh}{OH}{One-Hot}

% P
\newacronym{pdf}{PDF}{Probability Density Function}
\newacronym{pgm}{PGM}{Probabilistic Graphical Model}

% R
\newacronym{relu}{ReLU}{Rectified Linear Unit}
\newacronym{rdf}{RDF}{Resource Description Framework}
\newacronym{rgb}{RGB}{Red Green Blue}
\newacronym{rgc}{RGC}{Relational Graph Convolution}
\newacronym{rgat}{RGAT}{Relational Graph Attention Network}
\newacronym{rgcn}{RGCN}{Relational Graph Convolutional Network}
\newacronym{rhs}{R.H.S.}{Right Hand Side}
\newacronym{rnn}{RNN}{Recurrent Neural Network}
\newacronym{roc}{ROC}{Receiver Operating Characteristic}

% S
\newacronym{sgd}{SGD}{Stochastic Gradient Descent}
\newacronym{sg}{SG}{SkipGram}
\newacronym{simclr}{SimCLR}{Simple framework for Contrastive Learning of visual Representations}
\newacronym{st}{ST}{SkipThought}
\newacronym{sts}{STS}{Semantic Textual Similarity}

% T
\newacronym{termfreq}{TF}{Term Frequency}
\newacronym{tfidf}{TF-IDF}{Term Frequency-Inverse Document Frequency}
\newacronym[plural=TPPs, descriptionplural={Temporal Point Processes}]{tpp}{TPP}{Temporal Point Process}

% U

% V
\newacronym{vae}{VAE}{Variational Auto-Encoder}

% W
\newacronym{wirgat}{WIRGAT}{Within-Relation Graph Attention}
\newacronym{wl}{WL}{Weisfeiler-Lehman}

% X

% Y

% Z

%% file: preamble/style.tex
\ifappleformat
\else
\fi

\newcommand{\widetemplatefigurebegin}{\ifappleformat\begin{figure}[t]\else\begin{figure*}[t]\fi}
\newcommand{\widetemplatefigureend}{\ifappleformat\end{figure}\else\end{figure*}\fi}

\DeclareRobustCommand{\mathup}[1]{\begingroup\changegreek\mathrm{#1}\endgroup}
\DeclareRobustCommand{\mathbfup}[1]{\begingroup\changegreekbf\mathbf{#1}\endgroup}
\DeclareRobustCommand{\mathbit}[1]{\bm{\mathit{#1}}}

\DeclareMathAlphabet{\mathsfit}{\encodingdefault}{\sfdefault}{m}{sl}
\SetMathAlphabet{\mathsfit}{bold}{\encodingdefault}{\sfdefault}{bx}{n}
\newcommand{\tens}[1]{\bm{\mathsfit{#1}}}

\newcommand{\constantvector}{\bm}               % Use \va for vector-valued constant a.

\newcommand{\constantmatrix}{\bm}               % Use \mA for matrix-valued constant A.
\newcommand{\constantmatrixgreek}{\mathbit}

\newcommand{\randomscalar}{\textnormal}         % Use \ra for random variable a. Don't use m.
\newcommand{\randomscalargreek}{\mathup}

\newcommand{\randomvector}{\mathbf}             % Use \rva for vector-valued random variable a.
\newcommand{\randomvectorgreek}{\mathbfup}

\newcommand{\randommatrix}{\mathbf}             % Use \rmA for matrix-valued random variable A.
\newcommand{\randommatrixgreek}{\mathbfup}

\newcommand{\graphstyle}{\mathcal}              % Use \gA for graph A.

\newcommand{\tensorstyle}{\tens}                % Use \tA for tensor A.

\newcommand{\setstyle}{\mathbb}                % Use \sA for set A. Don't use E.

%% file: preamble/symbols.tex
\def\alphabet{a,b,c,d,e,f,g,h,i,j,k,l,m,n,o,p,q,r,s,t,u,v,w,x,y,z}
\def\Alphabet{A,B,C,D,E,F,G,H,I,J,K,L,M,M,O,P,Q,R,S,T,U,V,W,X,Y,Z}
\def\greekalphabet{alpha,beta,gamma,delta,epsilon,varepsilon,zeta,eta,theta,vartheta,iota,kappa,varkappa,lambda,mu,nu,xi,pi,varpi,rho,varrho,sigma,varsigma,tau,upsilon,phi,varphi,chi,psi,omega}
\def\GreekAlphabet{Gamma,Delta,Theta,Lambda,Xi,Pi,Sigma,Upsilon,Phi,Psi,Omega}

\makeatletter
\def\changegreek{\@for\next:=\greekalphabet
	\do{\expandafter\let\csname\next\expandafter\endcsname\csname\next up\endcsname}}
\def\changegreekbf{\@for\next:=\greekalphabet
	\do{\expandafter\def\csname\next\expandafter\endcsname\expandafter{%
			\expandafter\bm\expandafter{\csname\next up\endcsname}}}}
\makeatother

\foreach \x in \alphabet {
	\expandafter\xdef\csname v\x\endcsname{\noexpand\ensuremath{\noexpand\constantvector{\x}}}

	\expandafter\xdef\csname ev\x\endcsname{\noexpand\ensuremath{\noexpand\x}}

	\ifthenelse{\equal{\x}{m}}{}{
		\expandafter\xdef\csname r\x\endcsname{\noexpand\ensuremath{\noexpand\randomscalar{\x}}}
	}

	\expandafter\xdef\csname rv\x\endcsname{\noexpand\ensuremath{\noexpand\randomvector{\x}}}
}

\foreach \x in \greekalphabet {
	\expandafter\xdef\csname v\x\endcsname{\noexpand\ensuremath{\noexpand\constantvector{\csname \x\endcsname}}}

	\expandafter\xdef\csname ev\x\endcsname{\noexpand\ensuremath{\noexpand{\csname \x \endcsname}}}

	\expandafter\xdef\csname r\x\endcsname{\noexpand\ensuremath{\noexpand\randomscalargreek{\csname \x\endcsname}}}

	\expandafter\xdef\csname rv\x\endcsname{\noexpand\ensuremath{\noexpand\randomvectorgreek{\csname \x\endcsname}}}
}

\foreach \x in \Alphabet {
	\expandafter\xdef\csname m\x\endcsname{\noexpand\ensuremath{\noexpand\constantmatrix{\x}}}

	\expandafter\xdef\csname em\x\endcsname{\noexpand\ensuremath{\noexpand\x}}

	\expandafter\xdef\csname rm\x\endcsname{\noexpand\ensuremath{\noexpand\randommatrix{\x}}}

	\expandafter\xdef\csname t\x\endcsname{\noexpand\ensuremath{\noexpand\tensorstyle{\x}}}

	\expandafter\xdef\csname g\x\endcsname{\noexpand\ensuremath{\noexpand\graphstyle{\x}}}

	\ifthenelse{\equal{\x}{E}}{}{
		\expandafter\xdef\csname s\x\endcsname{\noexpand\ensuremath{\noexpand\setstyle{\x}}}
	}

}

\foreach \x in \GreekAlphabet {
	\expandafter\xdef\csname m\x\endcsname{\noexpand\ensuremath{\noexpand\constantmatrixgreek{\csname \x\endcsname}}}

	\expandafter\xdef\csname rm\x\endcsname{\noexpand\ensuremath{\noexpand\randommatrixgreek{\csname \x\endcsname}}}
}

%% file: preamble/math.tex
\DeclareRobustCommand{\norm}[1]{\ensuremath{\left\lVert#1\right\rVert}}

\DeclareRobustCommand{\KLD}[2]{\ensuremath{\textnormal{KLD}\left(#1\,\Vert\,#2\right)}}

\newcommand{\E}{\mathbb{E}}
\newcommand{\Ls}{\mathcal{L}}

\newcommand{\pdata}{p_{\rm{data}}}

%% file: paper_metadata.tex
\newcommand{\papertitle}{Scaling Properties of Continuous Diffusion Spoken Language Models}
\newcommand{\paperkeywords}{textless NLP, spoken language models, diffusion models, scaling laws}
\newcommand{\papercorrespondence}{Jason Ramapuram (jason@ramapuram.net); Eeshan Gunesh Dhekane (eeshan@apple.com); Amitis Shidani (amitis\_shidani@apple.com); Tatiana Likhomanenko (antares@apple.com)}
\newcommand{\paperabstract}{\textit{Speech-only} spoken language models (SLMs) lag behind text and text-speech models in performance, with recent discrete autoregressive (AR) SLMs indicating significant computational and data demands to match text models. Since discretizing continuous speech for AR creates bottlenecks, we explore whether \textit{continuous diffusion (CD) SLM} is more viable. To quantify the SLMs linguistic quality, we introduce the phoneme Jensen-Shannon divergence (pJSD) metric. Our analysis reveals CD SLMs, mirroring AR behavior, exhibit scaling laws for validation loss and pJSD, and show optimal token-to-parameter ratios decreasing as compute scales. However, for the latter, loss becomes insensitive to choice of data and model sizes, showing potential for fast inference. Scaling CD SLMs to 16B parameters with tens of millions of hours of conversational data enables generation of emotive, prosodic, multi-speaker, multilingual speech, though achieving long-form coherence remains a significant challenge.}
\newcommand{\paperworkdoneatapplesymbol}{$^\dagger$}
\newcommand{\papercoreadvisingsymbol}{$^\star$}

%% file: title_authors.tex
\title{\papertitle}

\ifappleformat
  \author{Jason Ramapuram$^{*\dagger}$}
  \author{Eeshan Gunesh Dhekane$^*$}
  \author{Amitis Shidani$^*$}
  \author{Dan Busbridge}
  \author{Bogdan Mazoure$^\dagger$}
  \author{Zijin Gu}
  \author{Russ Webb}
  \author{Tatiana Likhomanenko$^\star$}
  \author{Navdeep Jaitly$^{\dagger\star}$}

  \affiliation{Apple}

  \contribution[*]{Core contributor}
  \contribution[\star]{Core advising}
  \contribution[\dagger]{Work done while at Apple}

  \metadata[Correspondence]{\sffamily
  \papercorrespondence
  }
\else
  \author[affiliation={1}, equalcontribution]{Jason}{Ramapuram\paperworkdoneatapplesymbol}
  \author[affiliation={1}, equalcontribution]{Eeshan Gunesh}{Dhekane}
  \author[affiliation={1}, equalcontribution]{Amitis}{Shidani}
  \author[affiliation={1}]{Dan}{Busbridge}
  \author[affiliation={1}]{Bogdan}{Mazoure\paperworkdoneatapplesymbol}
  \author[affiliation={1}]{Zijin}{Gu}
  \author[affiliation={1}]{Russ}{Webb}
  \author[affiliation={1}]{Tatiana}{Likhomanenko\papercoreadvisingsymbol}
  \author[affiliation={1}]{Navdeep}{Jaitly\paperworkdoneatapplesymbol\papercoreadvisingsymbol}

  \address{
      $^1$ Apple
  }

  \email{jason@ramapuram.net, eeshan@apple.com, amitis\_shidani@apple.com, antares@apple.com}
  \keywords{\paperkeywords}
\fi

%% file: abstract.tex
\ifappleformat
  \abstract{\paperabstract}
\fi

%% file: body/03_01_introduction.tex
\section{Introduction}
\label{sec:introduction}
\ifappleformat
    \def\introfigurepanelwidth{0.49\textwidth}
\else
    \def\introfigurepanelwidth{0.47\textwidth}
\fi
\begin{figure}[!ht]
    \centering
    \begin{subfigure}[t]{\introfigurepanelwidth}
        \includegraphics[width=\textwidth]{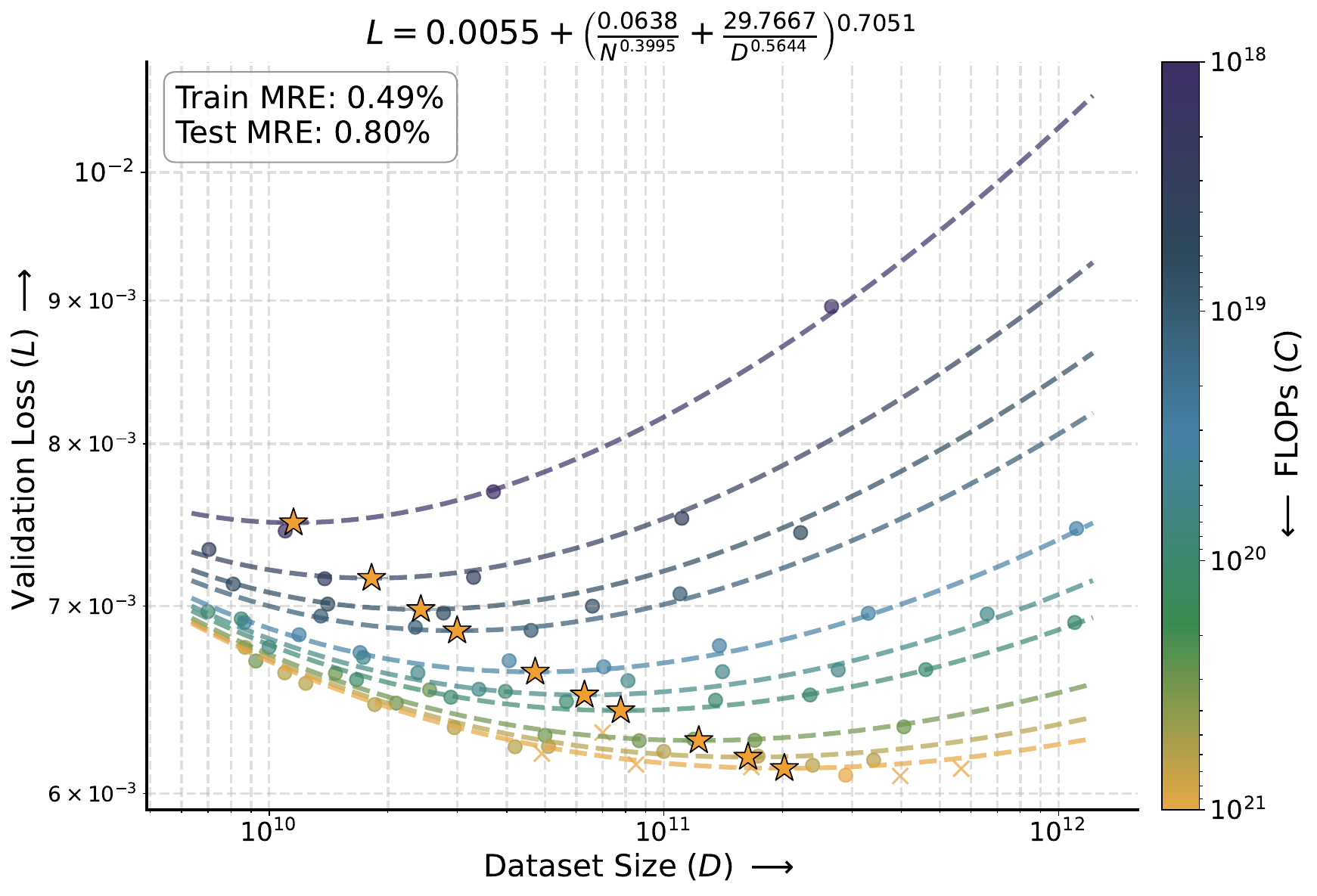}    
        % \caption*{(a)}
        \label{fig:related_work:header_figure:scaling_laws_fit:A}
    \end{subfigure}
    \hfill
    \begin{subfigure}[t]{\introfigurepanelwidth}
        \includegraphics[width=\textwidth]{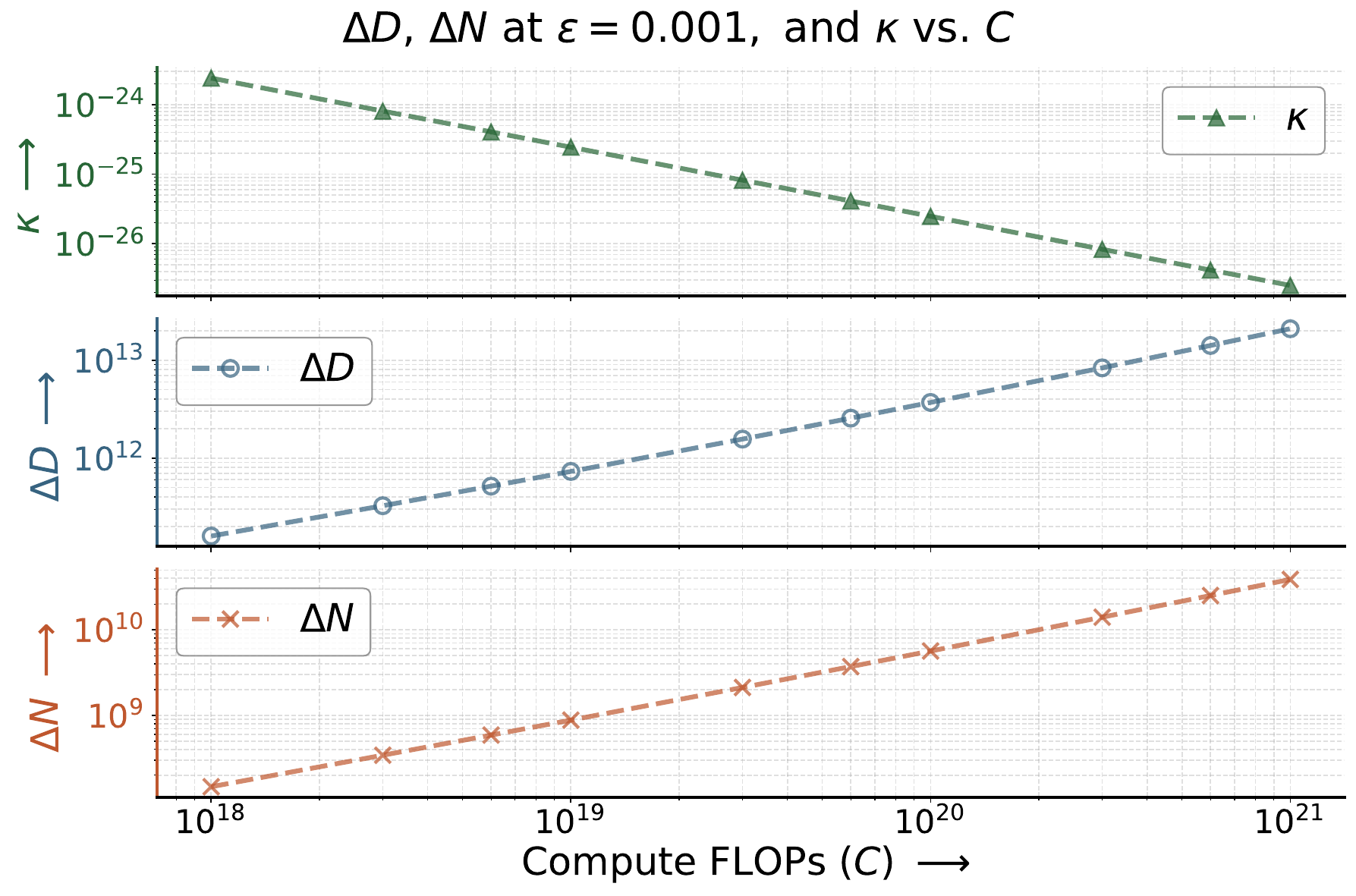}    
        % \caption*{(b)}
        \label{fig:related_work:header_figure:kappa_DeltaD_DeltaN_vs_C:B}
    \end{subfigure}
    \caption{
        \ifappleformat (Left) \else (Top) \fi
        Scaling law fit for validation loss. Training ($\bullet$) and testing ($\times$) points are shown alongside compute-optimal points~($\star$). 
        % Stars are annotated with the optimal token-to-parameter ratio $r^\ast$, which decreases at higher computes, reflecting increased data efficiency. 
        \ifappleformat (Right) \else (Bottom) \fi The curvature $\kappa$ of isoFLOPs at their optima decreases as compute increases: flattening corresponds to approximately $2$ orders of magnitude expansion in the range of model ($\Delta N$) and dataset ($\Delta D$) sizes yielding a loss within $\epsilon$ of the optimum $L^\ast$. Thus, higher computes allow near-optimal performance across a much wider variety of parameter-to-data allocations, opening up an efficient inference frontier.
    }
    \label{fig:related_work:header_figure}
\end{figure}
Building on recent advancements in self-supervised learning (SSL) for speech processing~\cite{mohamed2022self} and the emergence of phonetic structure within these learned representations~\cite{baevski2020wav2vec,choi2024self}, the research community has advanced \textit{textless NLP}, a field aimed at training spoken language models (SLMs) directly from speech without textual supervision.\footnote{Referred to as ``pure speech language models'' in~\cite{arora2025on}.}
% Using recent advancements in self-supervised learning (SSL) for speech processing~\cite{mohamed2022self} and emergence of phonetic structure in its learned representations~\cite{baevski2020wav2vec,choi2024self}, the research community has advanced the field of \textit{``textless NLP''}, aimed at training spoken language models (SLMs) directly from raw speech without textual supervision.\footnote{Referred to as ``pure speech language models'' in~\cite{arora2025on}.}
The prevailing methodology involves discretizing SSL representations into speech tokens and training autoregressive (AR) models on them.
While recent works have demonstrated significant progress using this paradigm~\cite{lakhotia2021generative,dunbar2021zero,kharitonov2022text,cuervo2024scaling,arora2025on}, current performance remains comparable to the linguistic proficiency of a three- to four-year-old child (according to metrics measuring lexical, syntactic and semantic proficiency), placing SLMs significantly behind the capabilities of state-of-the-art text-based and speech-text systems~\cite{cuervo2024scaling,maimon2025scaling,poli2025spidr}.

Bridging this performance gap requires addressing two fundamental challenges: determining optimal speech representations~\cite{poli2025spidr} and identifying the most effective modeling paradigm. 
The latter is particularly pressing as recent SLM scaling laws suggest that achieving LLM-level linguistic proficiency via AR modeling on discrete speech tokens \textit{could require orders of magnitude more compute~\cite{cuervo2024scaling}.}\footnote{Since \cite{cuervo2024scaling} fixed hyperparameters across all compute budgets, a practice known to scale suboptimally \cite{yang2021tensor,dey2025dont}, these computational requirements may be overestimated.}
% \footnote{Because~\cite{cuervo2024scaling} used fixed hyperparameters across all compute budgets and recent work~\cite{yang2021tensor,dey2025dont} showed that optimal hyperparameters under standard parametrization do not transfer across model scales, these scaling requirements may be overestimated.} 
This steep computational burden is likely compounded by the inherent challenges of raw speech: low information density, high acoustic and speaker variability, and a lack of semantically dense, curated data comparable to rich datasets like Wikipedia in text generative modeling. 
Consequently, extracting general knowledge from speech remains highly resource-intensive.
% Bridging this performance gap requires addressing several fundamental and unresolved challenges.
% First, the optimal speech representations for SLMs remains poorly understood, as highlighted by~\cite{poli2025spidr}.
% Second, the identification of the most effective modeling paradigm is still an open question. 
% The latter is particularly critical given that recent work on SLM scaling laws~\cite{cuervo2024scaling} predicts that \textit{achieving linguistic proficiency comparable to text-based large language models (LLMs) using AR modeling on discrete speech tokens may require several orders of magnitude}\footnote{Note that~\cite{cuervo2024scaling} used fixed hyperparameters for all compute budgets. 
% This may lead to overestimated scaling requirements given recent findings that optimal hyperparameters under standard parametrization do not transfer across model scales~\cite{yang2021tensor,dey2025dont}.} \textit{more compute}. 
% This inefficiency is likely compounded by the low information density of speech representations and the high diversity of acoustic and speaker variations compared to text. 
% Furthermore, raw speech data typically lacks the semantic density found in curated text corpora such as Wikipedia, making the extraction of general knowledge significantly more resource-intensive.

Inspired by recent successes of diffusion models in vision~\cite{velez2025image,croitoru2023diffusion,saharia2022photorealistic,rombach2022high} and text-speech AR models with speech continuous representations~\cite{meng2025autoregressive,rouard2025continuous}, \textit{we investigate continuous diffusion (CD) models as a potential alternative to discrete AR modeling for SLMs.}
Moreover, speech signals are continuous, even if the language they convey is discrete, thus looking beyond AR modeling on discrete tokens for SLMs is natural.

The potential of a modeling paradigm is defined by its scaling behavior. 
Given that ``languageness'', the ability to generate long-form coherent language, is the key metric in SLMs, we study how it behaves under scaling in CD SLMs. 
However, measuring languageness poses a challenge: unlike AR models, computing exact sequence log-likelihoods in continuous diffusion models is computationally prohibitive.
%unlike AR models that induce a likelihood factorization used to measure it, diffusion models lack direct access to a probability density.
% Diffusion models are generative models, yet unlike AR models, they lack direct access to the (unnormalized) probability density. 
To address this, 
% NB(jrp): bold always feels like yelling to me :) 
% \textbf{we first introduce a new metric, the phoneme Jensen-Shannon divergence (pJSD)} (\Cref{sec:phoneme-jsd}).
\emph{we first propose to use the phoneme Jensen-Shannon divergence (pJSD) metric} (\Cref{sec:phoneme-jsd}): it quantifies the model's ``languageness'' by computing the divergence between the phoneme $n$-grams distributions of generated and real speech. 
We then analyze the scaling behavior of CD SLMs, demonstrating that \textit{they exhibit trends similar to those of discrete AR SLMs~\cite{cuervo2024scaling}, and do not fundamentally alter the scaling trajectory, though exhibit some new practical scaling behavior not observed in prior work  (\Cref{sec:scaling-laws-for-continuous-diffusion-SLMs})}. Our results show that:
% compared to discrete AR approaches}:
%
\begin{itemize}
    \item (Known trend) Validation loss follows scaling laws (\Cref{fig:related_work:header_figure} (top) and~\Cref{subsec:scaling-laws-for-continuous-diffusion-SLMs:scaling-law-for-validation-loss}).
    \item (Known trend) The optimal token-to-parameter ratio is compute  dependent, decreasing as the compute budget scales (\Cref{fig:related_work:header_figure} (bottom) and~\Cref{subsec:scaling-laws-for-continuous-diffusion-SLMs:scaling-law-for-validation-loss}). 
    \item (New trend) Higher computes allow near-optimal performance across a much wider variety of parameter-to-data allocations (\Cref{fig:related_work:header_figure} (bottom) and~\Cref{subsec:scaling-laws-for-continuous-diffusion-SLMs:scaling-law-for-validation-loss}), opening up possibility for fast inference.
    % \item (Known trend) pJSD metric demonstrates that learned ``languageness'' scales predictably, mirroring discrete AR models (sBLIMP~\cite{dunbar2021zero}, sStoryCloze~\cite{hassid2023textually}) (\Cref{subsec:scaling-laws-for-continuous-diffusion-SLMs:scaling-law-for-evaluation-metrics}). Thus, pJSD provides a viable evaluation tool for generative models without direct access to (un)normalized probability densities.
    \item (Known trend) The pJSD metric demonstrates that learned ``languageness'' follows scaling laws, mirroring discrete AR models (sBLIMP~\cite{dunbar2021zero}, sStoryCloze~\cite{hassid2023textually}) (\Cref{subsec:scaling-laws-for-continuous-diffusion-SLMs:scaling-law-for-evaluation-metrics}). Thus, pJSD provides a viable sampling-based evaluation tool for generative models that do not offer the easily factorized likelihoods of autoregressive architectures.
    \item (New trend) Unlike prior work on AR SLMs~\cite{cuervo2024scaling}, we also analyze standard perceptual quality metrics. We find they \textbf{do not} exhibit scaling laws (this behavior is aligned with their poor correlation to human mean opinion scores~\cite{manku2025emergentttseval}). However, two out of four Meta Audiobox Aesthetics~\cite{tjandra2025meta} components (content enjoyment and content understanding) \textbf{do} scale predictably (\Cref{subsec:scaling-laws-for-continuous-diffusion-SLMs:scaling-law-for-evaluation-metrics}).
    \item (New trend) Metrics without scaling laws generally saturate near their real-data baselines. In contrast, for certain metrics, our best scaling fits suggest that real-data baselines remain unreachable at any compute budget (\Cref{subsec:scaling-laws-for-continuous-diffusion-SLMs:scaling-law-for-evaluation-metrics}).
\end{itemize}
Finally, we scale CD SLMs to a 16B parameter model trained on tens of millions of hours of conversational speech (\Cref{sec:aux_cond}). 
While at that scale our model generates multi-speaker multilingual conversations with rich emotions and prosody, achieving long-form linguistic coherence remains a significant challenge.
\textit{This shortfall suggests that given current compute budgets and available speech data, further scaling of SLMs is impractical unless a new speech representation or modeling paradigm emerge, or we pivot to text-speech models~\cite{maimon2025scaling,arora2025on}}.\footnote{
We focus exclusively on pretraining, where foundational representations emerge. While post-training is highly effective for steering and refining behavior, there is no evidence it can instill basic linguistic coherence if the base model lacks it.}

%% file: body/03_02_related_work.tex
\section{Related Work}
\label{sec:related_work}

\noindent\textbf{Speech Representations}~~
Since the inception of SLMs~\cite{lakhotia2021generative}, speech generative modeling has largely converged on AR modeling of discrete speech tokens~\cite{mousavi2025discrete,arora2025on}, a paradigm favored for its compatibility with pretrained text LLMs. 
To mitigate bottlenecks arisen from discretization of continuous speech signals, recent works have explored AR modeling on continuous representations for both text-to-speech~\cite{meng2025autoregressive} (log-mel filterbanks) and text-speech LLMs~\cite{rouard2025continuous}. 
Yet, these approaches often face training convergence issues, necessitating auxiliary loss functions or variational components. 
Instead we apply continuous diffusion directly to log-mel filterbanks: a native fit for continuous data.
% In contrast, diffusion models are natively designed for continuous data distributions. 
% In this work we focus on continuous diffusion on continuous log-Mel filterbanks.
% Thus, we investigate diffusion modeling directly on continuous speech representations, offering a more natural fit for the modality without the complexities % required to stabilize continuous AR models.  % NB(jrp): AR is very stable if done right :) 
% ND(tata): good luck! if you can make it - happy to see your paper asap on getting it to work!
% of continuous AR SLMs.

\noindent\textbf{Diffusion Models}~~
% Diffusion models~\cite{ho2020ddpm,DBLP:journals/corr/abs-1907-05600,DBLP:conf/iclr/0011SKKEP21} have emerged as a dominant generative paradigm, achieving state-of-the-art results in image synthesis~\cite{dhariwal2021diffusion,rombach2022high,saharia2022photorealistic,DBLP:conf/icml/EsserKBEMSLLSBP24}. In video generation, diffusion transformers operating on spacetime latent patches have enabled high-fidelity synthesis, as demonstrated by Sora~\cite{openai2024sora,peng2025open} and Movie Gen~\cite{polyak2024movie}. 
% Diffusion has also been explored for language modeling:~\cite{li2022diffusion} introduced Diffusion-LM for controllable text generation, and more recent work such as 
%MDLM~\cite{sahoo2024simple} 
% LLaDa\cite{DBLP:journals/corr/abs-2502-09992} has shown that discrete diffusion models can approach autoregressive performance on text benchmarks.
Diffusion models~\cite{ho2020ddpm,DBLP:journals/corr/abs-1907-05600,DBLP:conf/iclr/0011SKKEP21} have emerged as a dominant generative paradigm, achieving state-of-the-art results in image~\cite{dhariwal2021diffusion,rombach2022high,saharia2022photorealistic,DBLP:conf/icml/EsserKBEMSLLSBP24} and video~\cite{openai2024sora,peng2025open,polyak2024movie} generation, and approaching AR models in language~\cite{li2022diffusion,DBLP:journals/corr/abs-2502-09992}.
Recently, diffusion models have been applied to audio and music generation~\cite{DBLP:conf/icml/EvansCTHP24,DBLP:conf/ismir/EvansPCZTP24}.
% In the speech domain, DiffWave~\cite{kong2021diffwave} was among the first to apply diffusion to raw waveform generation from a Mel-spectrogram as a neural vocoder, demonstrating competitive quality with AR and GAN-based vocoders. 
% Subsequent work extended diffusion to text-to-speech (TTS): Grad-TTS~\cite{popov2021grad} formulated Mel-spectrogram synthesis as a stochastic differential equation, while Diff-TTS~\cite{jeong2021diff} and ProDiff~\cite{huang2022prodiff} improved inference speed through knowledge distillation and progressive denoising, respectively. More recently, diffusion-based TTS systems have been scaled with transformer backbones~\cite{DBLP:conf/iccv/PeeblesX23} and latent representations, with models such as NaturalSpeech2~\cite{shen2023naturalspeech} and VoiceBox~\cite{le2023voicebox} achieving near-human quality by operating in latent codec spaces with classifier-free guidance~\cite{ho2022classifierfree}. 
In speech domain, diffusion was first applied to neural vocoders~\cite{kong2021diffwave}, showing competitive quality with AR and GAN-based models. 
Subsequent work extended diffusion to text-to-speech (TTS)~\cite{popov2021grad,jeong2021diff,huang2022prodiff,DBLP:conf/iccv/PeeblesX23,shen2023naturalspeech,le2023voicebox}, achieving near-human quality by operating in latent codec spaces with classifier-free guidance~\cite{ho2022classifierfree}.
E2 TTS~\cite{eskimez2024e2} further simplified the pipeline by directly generating mel-spectrograms from text in a fully end-to-end diffusion framework. 
Recently,~\cite{zhou2025diffa} introduced DIFFA, the first
diffusion-based speech LLM designed to perform spoken language understanding, building on top of a frozen discrete diffusion-based LM.
However, all of these approaches rely on explicit text conditioning or text-pretrained models.
To the best of our knowledge, our work is the first to apply continuous diffusion models to SLM \textit{without any text supervision}, investigating their scaling properties.
% in this textless setting.

\noindent\textbf{Scaling Laws}~~
% The systematic study of neural scaling laws was initiated by~\cite{kaplan2020scaling}, which demonstrated that the cross-entropy loss of AR language models (LMs) follows predictable power-law relationships with respect to model size, dataset size, and training compute budget, and that these relationships hold over several orders of magnitude. 
The systematic study of neural scaling laws was initiated by~\cite{kaplan2020scaling}, showing the cross-entropy loss of AR LMs follows power-law relationships with respect to model size, data size, and training compute budget, holding over several orders of magnitude. 
\cite{DBLP:journals/corr/abs-2203-15556} refined these findings by showing that prior work had significantly undertrained or overtrained models relative to their size: for a fixed compute budget, model parameters and training tokens should be scaled in roughly equal proportion, a result that shifted practical training recipes toward compute optimal. 
% Beyond standard training, \cite{DBLP:conf/icml/BusbridgeSWRLW25} established scaling laws for knowledge distillation, showing that the optimal allocation of compute between teacher and student models also follows predictable power-law relationships. 
% In computer vision, \cite{zhai2022scaling} established scaling laws for vision transformers on classification tasks, finding similar power-law trends between error rate and compute, though with different exponents than language. 
Beyond LM training, \cite{DBLP:conf/icml/BusbridgeSWRLW25} established scaling laws for knowledge distillation, and \cite{zhai2022scaling}~-- for vision transformers on classification tasks.
% For multimodal models, \cite{cherti2023reproducible} studied scaling behaviour of CLIP-style contrastive learning \cite{radford2021learning} across model and data size, observing that both axes contribute to downstream zero-shot performance following predictable trends. \cite{aghajanyan2023scaling} further investigated scaling laws for causal masked multimodal models, finding that jointly training on text and images, or text and speech can improve scaling efficiency for both modalities. 
For multimodal models, \cite{cherti2023reproducible} observed scaling behavior of CLIP-style contrastive learning \cite{radford2021learning}, and \cite{aghajanyan2023scaling} further investigated causal masked multimodal models, finding that jointly training on text and images, or text and speech improves scaling efficiency for both modalities. 
For masked discrete diffusion models, \cite{DBLP:journals/corr/abs-2512-10858} studied the scaling behavior of LMs, finding they follow similar loss scaling trends as AR models but with a constant efficiency gap. \cite{bethune2026designspacetrimodalmasked} provided the first scaling laws for multimodal (image, audio and text) case and showed the token-per-parameter ratio decreases as compute grows.
For speech, \cite{cuervo2024scaling} established the first scaling laws for SLMs using AR modeling on discrete tokens, while \cite{maimon2025scaling} extended this analysis to interleaved text-speech models.
Our work complements these efforts by providing the first scaling law analysis for \textit{continuous diffusion SLMs}.

%% file: body/03_03_experiments.tex
\section{Continuous Diffusion SLMs}
\label{sec:continuous-diffusion-SLMs}
% Since scaling laws are sensitive to the choice of training data~\cite{kaplan2020scaling}, data representation~\cite{liu2025superposition}, and model architecture~\cite{kaplan2020scaling}, we first define these components to establish the precise conditions under which we analyze the scaling behavior of continuous diffusion SLMs.

\subsection{Data}
We use a large-scale conversational speech dataset, dubbed SpeechCrawl, collected from publicly accessible sources. 
This dataset was specifically curated to be diverse, conversational, multilingual, and multi-speaker. 
Audio samples average approximately 30 minutes in duration, with roughly 60\% consisting of English speech. 
As SpeechCrawl lacks metadata, we employ the WhisperX~\cite{bain2023whisperx} pipeline, utilizing Whisper large-v3 multilingual model~\cite{radford2023robust}, to determine the percentage of English speech in each sample. Subsequently, we filter the dataset to retain only those audio samples exceeding 5 minutes in duration where English comprises at least 99\% of the speech content. The resulting filtered dataset has 7 million hours of speech.

\subsection{Speech Representation}
\label{sec:speech-repres}
We construct our diffusion SLMs using log-mel filterbanks, a choice motivated by their distinct advantages over data-driven approaches~\cite{bai2024dmel}. 
Unlike neural-based representations, which often introduce compression artifacts and limit generalization, log-mel filterbanks provide a physics-based, interpretable representation that preserves both semantic and acoustic details with minimal information loss. 
Additionally, this representation is model-agnostic, decoupling the generation process from specific encoder-decoder architectures and allowing for waveform reconstruction via any compatible vocoder.
Finally, log-mel filterbanks offer proven reliability and domain agnostic performance across diverse acoustic environments~\cite{tseng2025probing}.

We resample all SpeechCrawl audio to 24kHz and extract 80-dimensional log-mel filterbanks (50ms window, 12.5ms hop), resulting in an 80 Hz frame rate. 
To contextualize this density, consider standard heuristics: text-based LLMs typically average \textit{four} tokens per \textit{three} words, while conversational speech averages about \textit{three} words per second. 
Consequently, one second of speech corresponds to \textit{four} text tokens versus 80 spectral frames. This represents a $20\times$ increase in sequence length (or token number) for equivalent semantic content.
% {\color{red} TBD add what Santiago's paper did on tokenization and how this scales. - he was using 25Hz, so 3.2x smaller -> scaling laws should be different by 6.25x with LLMs -> 11/14 tokens per sec if we do unigram}

\subsection{Continuous Diffusion (CD) Model}
\label{sec:model}

\begin{figure}[!t]
  \centering
  \ifappleformat
    \includegraphics[width=0.70\linewidth]{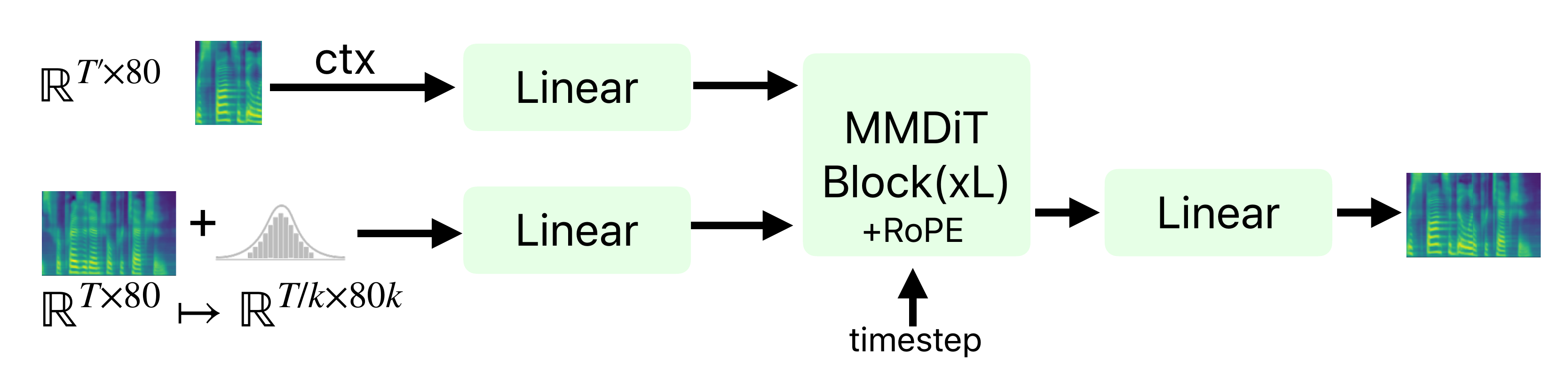}
  \else
    \includegraphics[width=0.88\linewidth]{plotting/audio_diffusion_model_diag.pdf}
  \fi
  \caption{Continuous diffusion SLM architecture.}
  \label{fig:main_model}
\end{figure}

Continuous diffusion models~\cite{ho2020ddpm,DBLP:journals/corr/abs-1907-05600,DBLP:conf/iclr/0011SKKEP21} define a generative process by learning to reverse a fixed forward noising process. Given a data sample $x_0 \sim \pdata$, the forward process produces a sequence of increasingly noisy latents $\{x_t\}_{t=0}^{T}$ according to 
\begin{equation}
    q(x_t | x_0) = \mathcal{N}(x_t; \sqrt{\bar{\alpha}_t} x_0, (1 - \bar{\alpha}_t) \mathbf{I}),
\end{equation} 
where $\bar{\alpha}_t = \prod_{s=1}^{t} \alpha_s$ defines the cumulative noise schedule and $\alpha_t = 1 - \beta_t$ with $\beta_t$ controlling the noise added at each step. As $t \to T$, the distribution $q(x_T)$ approaches an isotropic Gaussian prior $\mathcal{N}(0, \mathbf{I})$. The reverse process learns to denoise by parameterizing a neural network $\epsilon_\theta(x_t, t)$ to predict the noise $\epsilon$ added to form~$x_t$. 
Following~\cite{DBLP:conf/iclr/SalimansH22,DBLP:journals/corr/abs-2511-13720}, we instead parameterize a neural network $v_\theta(x_t,t)$ to predict the velocity $v_t = \sqrt{\bar{\alpha}_t} \epsilon - \sqrt{1 - \bar{\alpha}_t} x_0$, which interpolates between predicting noise and signal. 
% We adopt velocity $v_t$ prediction with $x_0$-reconstruction in this work, as recent findings~\cite{DBLP:journals/corr/abs-2511-13720} suggest that predicting the clean signal is advantageous since data lies in a low-dimensional subspace while noise is full-rank. 
We minimize a min-SNR~\cite{hang2023efficient_minsnr} weighted denoising loss
\begin{equation}
\Ls = \E_{x_0, \epsilon, t} \left[ \min\left(\text{SNR}(t), \psi\right) \cdot \norm{v_\theta(x_t, t) - v_t}^2 \right],
\end{equation}
where $\text{SNR}(t) = \bar{\alpha}_t / (1 - \bar{\alpha}_t)$ is the signal-to-noise ratio at timestep~$t$ and $\psi$ is a truncation constant.
This reweighting addresses the imbalanced loss contributions across timesteps, improving training efficiency~\cite{hang2023efficient_minsnr}.
  
We adopt the \gls{mmdit} architecture~\cite{DBLP:conf/icml/EsserKBEMSLLSBP24}. \gls{mmdit} extends the standard \gls{dit}~\cite{DBLP:conf/iccv/PeeblesX23} framework, generalizing it from class-conditional image generation to support variable-length text conditioning. 
For our CD SLM, we adapt \gls{mmdit} by replacing the original text and image streams with two streams of log-mel filterbanks: one representing the audio context and the other representing the target continuation to be generated.

Given an original mono audio waveform, $x \in \mathbb{R}^{S \times 1}$, we convert the signal to 80 log-mel filterbanks, $m \in \mathbb{R}^{S' \times 80}$. 
We chunk $m$ into two segments: the context $m_{\text{ctx}} \in \mathbb{R}^{T' \times 80}$ and the signal we want to generate (the continuation), $m_{\text{gen}} \in \mathbb{R}^{T \times 80}$. 
Our model, highlighted in Figure~\ref{fig:main_model}, then proceeds to add Gaussian noise to $m_{\text{gen}}$ and project both signals to $d_{\text{emb}}$, the model embedding dimension, before relaying the signals to the underlying \gls{mmdit} model. 
\gls{mmdit} ensures that both streams (context and continuation) have independent pathways for all components such as AdaLN-zero~\cite{DBLP:conf/iccv/PeeblesX23} normalization layers, MLPs, projections, etc. in the transformer. 
The only interaction between the context and continuation streams takes place inside attention, where $Q$, $K$, and $V$ for each stream are concatenated and passed into a full bidirectional self-attention layer. 
This process is repeated for $L$ layers and the final data stream from continuation plus noise stream is extracted and passed into a diffusion loss. 
In \Cref{sec:scaling-laws-for-continuous-diffusion-SLMs}, we use 10s for $m_{\text{ctx}}$ and 30s for $m_{\text{gen}}$, and scale the model size by keeping $d_{\text{emb}}/L=128$~\cite{DBLP:conf/icml/BusbridgeSWRLW25}.

\noindent{\bf Classifier-Free Guidance (CFG)~\cite{ho2022classifierfree}}~~strengthens conditioning without a separate classifier. 
Standard CFG randomly drops the conditioning signal during training to jointly learn conditional and unconditional models. 
We found that this explicit dropping is unnecessary, and avoiding unconditional training steps saves substantial FLOPs, dedicating the full computational budget to the challenging conditional distribution \cite{DBLP:conf/nips/KarrasAKLAL24}.

At inference, we encode a signal of zeros to represent audio silence. This naturally serves as the unconditional signal $v_\theta(x_t, t, \varnothing)$. This approach aligns with projective composition~\cite{DBLP:conf/icml/BradleyNBTS25}, where evaluating score combinations against an empty background effectively isolates conditional features. Our zeroed speech signal provides this exact empty background, allowing the guidance equation to cleanly amplify the conditional score delta. The guided prediction is 
\begin{equation}
\tilde{v}_\theta(x_t, t, c) = v_\theta(x_t, t, \varnothing) + w \cdot \left( v_\theta(x_t, t, c) - v_\theta(x_t, t, \varnothing) \right),
\end{equation}
where $w > 1$ amplifies the conditioning signal. Since unguided generation ($w=1$) produced poor samples, we explore \textit{weak} ($w=2$) and \textit{strong} ($w=4$) CFG scales.

\subsection{Languageness Metric: Phoneme Jensen-Shannon Divergence (pJSD)}
\label{sec:phoneme-jsd}

Prior work on SLMs~\cite{lakhotia2021generative,dunbar2021zero,kharitonov2022text,borsos2023audiolm,poli2025spidr} assesses linguistic capabilities by computing the sWUGGY~\cite{dunbar2021zero} (lexical), sBLIMP~\cite{dunbar2021zero} (syntactic), and sStoryCloze~\cite{hassid2023textually} (semantic) metrics.
Fundamentally, these metrics evaluate whether the model assigns a higher probability to a linguistically correct sequence of discrete speech tokens compared to an incorrect counterpart (e.g., comparing the probability of a grammatically correct sentence against the same sentence containing a grammatical error). These evaluations rely on carefully curated datasets of paired words or sentences.

Since diffusion models lack easy access to the probability density of a data sample, we instead propose to measure the difference between \textit{empirical distributions of phoneme $n-$grams} of real data and generated data. 
In order to get generated data, we first sample log-mel filterbanks from the diffusion model and then synthesize raw speech by passing these log-mel filterbanks through a vocoder.\footnote{For simplicity we use of-the-shelf HifiGAN~\cite{kong2020hifi} vocoder~\url{https://github.com/kan-bayashi/ParallelWaveGAN/blob/master/egs/libritts/voc1/conf/hifigan.v1.yaml}.} 
% Since metrics such as sWUGGY, sBLIMP, and sStoryCloze rely on model-based scoring functions to rank specific pair of words, applying them to diffusion models would require developing equivalent scoring mechanisms. 
% Instead, we propose assessing diffusion models directly via their generated samples comparing the $n$-gram distributions. 
Then, given a waveform $x$ (real or generated), we extract a phoneme token sequence using a universal phoneme recognizer~\cite{li2020universalphone,allosaurus_github}.
% \footnote{We use the \texttt{allosaurus} library~\cite{allosaurus_github}.} 
Let this sequence be denoted as $\pi(x) = (p_1, p_2, \dots, p_L)$. 
For an integer $n \ge 1$, define the $i$-th contiguous phoneme $n$-gram as
\begin{equation}
g_i^{(n)}(x) := (p_i, p_{i+1}, \dots, p_{i+n-1}), \qquad i = 1, \dots, L-n+1.
\end{equation}
Let $C_{\mathcal{S}}^{(n)}(g)$ denote the total number of occurrences of $n$-gram $g$ aggregated over a corpus $\mathcal{S}$. Let $\Omega^{(n)}$ be the union support of $n$-grams observed in the generated set $\mathcal{G}$ or real data set $\mathcal{R}$. 
We then compute empirical distribution of the n-grams $g \in \Omega^{(n)}$ as
\begin{equation}
p_{\mathcal{S}}^{(n)}(g) := \frac{C_{\mathcal{S}}^{(n)}(g)}{Z_{\mathcal{S}}^{(n)}}, \qquad \,\, \,\,\, Z_{\mathcal{S}}^{(n)} := \sum_{g \in \Omega^{(n)}} C_{\mathcal{S}}^{(n)}(g).
\end{equation}
Let $m^{(n)} := \frac{1}{2}(p_{\mathcal{G}}^{(n)} + p_{\mathcal{R}}^{(n)})$. 
We report the phoneme Jensen-Shannon divergence (pJSD) \cite{lin1991divergence}, defined as
\begin{equation}
    \begin{aligned}
        \textnormal{pJSD}_n(\mathcal{G},\mathcal{R}) := {} & \frac{1}{2}\KLD{p_{\mathcal{G}}^{(n)}}{m^{(n)}} \\
        & + \frac{1}{2}\KLD{p_{\mathcal{R}}^{(n)}}{m^{(n)}}.
    \end{aligned}
\end{equation}
Lower values indicate closer agreement between empirical distributions of phoneme $n$-grams.\footnote{
% While pJSD measures the divergence between the empirical distributions of phoneme $n-$grams in real and generated data—a metric inherently limited by finite sample sizes—sWUGGY, sBLIMP, and sStoryCloze operate on a fundamentally different principle. Rather than being special cases of pJSD, these metrics are restricted to targeted pairwise comparisons of model likelihoods on curated sequences.
pJSD evaluates how closely the empirical distributions of phoneme $n-$grams match between real and generated data, though its accuracy is bounded by the finite size of the sampled data. Conversely, sWUGGY, sBLIMP, and sStoryCloze are not distributional metrics (nor are they a subset of pJSD); they are discriminative tasks limited strictly to pairwise comparisons of carefully curated correct and incorrect sequences.
} 
% [Eeshan]: I am commenting this line about n = 3 as we are reporting n = 1 and n = 5 now. 
%   For the fuller version of the paper, all n's to plot and if time permits, n = 7, 10.
% Unless otherwise stated, we report results for $n=3$. 

% [Resolved] Addition of reasons for avoiding ASR + LM pipeline. 
% Note that we do not use automatic speech recognition (ASR)-based evaluations for gauging the ``languageness'' learned by the model. 
% This is because our models demonstrate the capability of learning word bigrams/trigrams, thereby indicating the onset of learning languageness from speech.
% While evaluating these models, we observe that carrying out ASR followed by analyzing the transcripts (for instance, computing log-likelihood under a pre-trained language model) yields metric values that are highly noisy, thereby limiting their utility.
% However, exploring and developing ASR-based techniques for models at the onset of learning a language is an interesting direction, which we leave for future work. 
%
% NB(jrp): tightened

An advanced approach to capture the ``languageness'' learned in a model would be to measure the perplexity of the generated language by passing the audios synthesized by that model through an automatic speech recognition (ASR) system and a strong language model (LM). 
However, given that current SLMs perform at the level of a three- to four-year-old child, transcribing such language is challenging for ASR models, yielding a metric with high variance. As SLMs progress and learn to generate long-form, coherent language, this cascaded evaluation method—which is standard in other speech domains—should be adopted in the future instead of pJSD.

\subsection{Automatic Speech Perceptual Quality Metrics}
\label{sec:perception-metrics}
Most prior work on SLMs~\cite{lakhotia2021generative,dunbar2021zero,kharitonov2022text,borsos2023audiolm,poli2025spidr} focuses only on measuring linguistic capabilities, neglecting non-linguistic or paralinguistic dimensions. One of the contributions of our work is to 
% However, for the first time we 
assess generated speech not only for linguistic capabilities but also for perceptual quality.
We analyze automatic mean opinion scores (MOS) and Meta Audiobox Aesthetics~\cite{tjandra2025audiobox_aesthetics}.
\noindent\textbf{DNSMOS P.808~\cite{reddy2021dnsmos}} is a non-intrusive neural MOS predictor. The P.808 variant is designed to match ratings collected under an ITU-T P.808-style crowdsourcing protocol \cite{naderi2020p808}.
\noindent\textbf{DNSMOS overall (P.835)~\cite{reddy2022dnsmos_p835}} predicts perceptual quality dimensions aligned with ITU-T P.835 \cite{reddy2022dnsmos_p835,naderi2020p835}.
\noindent\textbf{NISQA MOS~\cite{mittag2021nisqa}} is a non-intrusive speech quality assessment model trained to predict MOS and related dimensions. 
\noindent\textbf{Meta AudioBox Aesthetics} provide learned no-reference predictors $f_k(x)$ for subjective axes $k$ such as content enjoyment, content understanding, production quality, and production complexity. 
For all predictors we report the mean score.
We exclude other non-linguistic or paralinguistic evaluations, as we posit these attributes are better evaluated after the post-training stage rather than at the pretraining stage, as they may be task-dependent.

%%%%%%%%%%%%%%%%%%%%%%%%%%%%%%%%%%%%%%%%%%%%%%%%%%%%%%%%%%%%%%%%%%%%%%%%%%%%%%%%
%%%%%%%%%%%%%%%%%%%%%%%%%%%%%% START: Scaling Laws for Continuous Diffusion SLMs
%%%%%%%%%%%%%%%%%%%%%%%%%%%%%%%%%%%%%%%%%%%%%%%%%%%%%%%%%%%%%%%%%%%%%%%%%%%%%%%%

\section{Scaling Laws for Continuous Diffusion SLMs}
\label{sec:scaling-laws-for-continuous-diffusion-SLMs}

\ifappleformat\else
    \input{body/isoflop_summary_figure}
\fi

Neural scaling laws characterize predictable performance improvements with respect to model size $N$, dataset size~$D$, and compute $C$~\cite{kaplan2020scaling}. For a fixed compute budget $C$, there exists an optimal allocation $(N^{\star}, D^{\star})$ that minimizes loss~\cite{DBLP:journals/corr/abs-2203-15556}. Following~\cite{kaplan2020scaling,gordon2021data}, we formalize this using the parametric scaling surface
\begin{align}
    L(N, D) = E + \left(\frac{A}{N^{\alpha}} + \frac{B}{D^{\beta}}\right)^\gamma\,,
    \label{eq:chinchilla_scaling}
\end{align}
where $E$ represents the irreducible entropy of the data, and the subsequent terms model approximation and estimation errors. 
While many empirical studies simplify this by setting $\gamma=1$~\cite{DBLP:journals/corr/abs-2203-15556,muennighoff2023scaling,choshen2025a}, we retain the outer $\gamma$ exponent for our CD SLMs as it significantly improves empirical stability~\cite{DBLP:conf/icml/BusbridgeSWRLW25}. 
To identify compute-optimal configurations, we employ the IsoFLOP methodology~\cite{kaplan2020scaling,DBLP:journals/corr/abs-2203-15556}, sweeping $(N, D)$ pairs under the constraint $C \approx 6ND$. Valid scaling dictates what we term \textit{expected isoFLOP behavior}: curves must exhibit a clear optimum ($\cup$-shaped for loss, $\cap$-shaped for quality metrics) alongside monotonic improvement as total compute increases.
Translating these loss-based scaling laws to downstream task metrics poses a distinct challenge. 
In our initial trials, direct-fitting approaches~\cite{DBLP:conf/iclr/Isik00PVK25,DBLP:journals/corr/abs-2512-08894} that bypass the loss entirely yielded poor mean relative error (MRE). 
Conversely, standard two-stage pipelines (fitting the loss, then mapping to the metric\footnote{In the infinite data regime, training loss $\approx$ validation loss.})
suffer from error accumulation~\cite{DBLP:journals/corr/abs-2107-03374,DBLP:journals/corr/abs-2412-04403}, leading to poor MRE.
To resolve this, \textbf{we introduce a fused two-stage approach}, jointly optimizing the parameters of both the scaling law and the downstream metric mapping, as detailed in Section~\ref{subsec:scaling-laws-for-continuous-diffusion-SLMs:scaling-law-for-evaluation-metrics}.
We analyze scaling behavior across ten compute budgets $C\in\{10^{18}, {3\cdot10^{18}}, {6\cdot10^{18}}, 10^{19}, {3\cdot10^{19}}, {6\cdot10^{19}}, 10^{20},$ ${3\cdot10^{20}}, {6\cdot10^{20}}, 10^{21}\}$, and model sizes ranging from ${\sim}0.6$M ($1$ layer) to ${\sim}11.5$B ($27$ layers) parameters. 
% For each budget, we sweep over $(N, D)$ pairs satisfying $C \approx 6ND$~\cite{DBLP:journals/corr/abs-2203-15556}, with model sizes ranging from ${\sim}0.6$M ($1$ layer) to ${\sim}11.5$B ($27$ layers) parameters. 
% Each configuration is trained with at least three seeds and evaluated on the metrics described in~\Cref{sec:phoneme-jsd,sec:perception-metrics}. 
Throughout the paper we report mean and standard deviation $\sigma$ for losses and evaluation metrics from~\Cref{sec:phoneme-jsd,sec:perception-metrics} across at least three training seeds.
All our model-training runs utilize scaled values of relevant hyperparameters as guided by muP~\cite{yang2021tensor} and completeP~\cite{dey2025dont} methods.
For this, we first perform hyperparameter tuning on a moderately sized base model of $\sim36\textrm{M}$ (4 layers) parameters, which is trained sufficiently for approximately $20\mathrm{k}$ steps. 
We tune learning rate in $\left\{1\mathrm{e}{-4}, 3\mathrm{e}{-4}, 4\mathrm{e}{-4}, 1\mathrm{e}{-3}, 3\mathrm{e}{-3}, 1\mathrm{e}{-2}\right\}$ and weight decay in $\left\{0.001, 0.003, 0.007, 0.01, 0.03, 0.07, 0.1, 0.2\right\}$ as these two are found to be the crucial hyperparameters.\footnote{
    The number of inference steps and the noise scheduler type is fixed due to compute limitations.
}
The best combination of learning rate $0.001$ and weight decay $0.03$ for the base model is used along with muP and completeP scaling to set hyperparameters across all our runs.

%\noindent This section describes the methodology for understanding scaling behavior of the validation loss and downstream metrics, lists our results, and their implications. 
%Towards this, we consider compute budgets $C$, where $C\in\left\{1\mathrm{e}18, 3\mathrm{e}18, 6\mathrm{e}18, 1\mathrm{e}19, 3\mathrm{e}19, 6\mathrm{e}19, 1\mathrm{e}20, 3\mathrm{e}20, 6\mathrm{e}20, 1\mathrm{e}21\right\}$.
% \footnote{\noindent We additionally consider the compute budget of $1\mathrm{e}21$ for understanding the scaling properties better.}.
%For each compute budget $C$, we train models with different model sizes $N$ and appropriately chosen dataset sizes $D$ such that the approximation $C \approx 6\cdot N\cdot D$ holds~\cite{kaplan2020scaling,DBLP:journals/corr/abs-2203-15556}. 
%For each such $(N, D)$ setting, we train multiple seeds for each setting so as to get better estimates of tracked metrics. 
% {\textcolor{red}{[Eeshan] Make sure to add or cref to a section that talks about the details of model training.}}
%Each of these trained models is then evaluated for the downstream metrics described in~\Cref{subsec:continuous-diffusion-SLMs:evaluation-metrics}. 
%These metric values and the validation loss of the models are cataloged for scaling law fitting.

\subsection{IsoFLOP Analysis}
\label{subsec:scaling-laws-for-continuous-diffusion-SLMs:IsoFLOP-analysis}
\noindent We analyze isoFLOP curves for validation loss and all evaluation metrics by plotting a curve of a particular metric versus the dataset size ($D$)
% NB(jrp): doesn't bring much value, takes space.
% \footnote{Equivalently, one could also plot the isoFLOPs with model size $N$ on the $x$-axis.} 
 for each compute level.
% For all isoFLOPs, and for all baselines wherever applicable, we compute the metric values on samples in the runs corresponding to those settings and compute the mean and standard deviation (denoted by $\sigma$) for the tracked metric values. 
% Ideally, the expected isoFLOP behavior is as follows: 
% \begin{enumerate}
%     \item At any compute budget, overparametrized but undertrained models as well as underparametrized but overtrained models lead to suboptimal metric values, and between these extremes, there is an optimal point where the metric value is best. 
%     Consequently, at any compute budget, the isoFLOPs of ``the-lower-the-better'' metrics are $\cup$-shaped and those of ``the-higher-the-better'' metrics are $\cap$-shaped curves.
%     \item Further, the optimal metric value per compute improves with more compute. 
% \end{enumerate}
% We observe that:

The key takeaways~are:
\begin{enumerate}
    \item \emph{Validation loss exhibits the expected isoFLOP behavior}; see \Cref{fig:subsec:scaling-laws-for-continuous-diffusion-SLMs:IsoFLOP-analysis:000003_isoFLOPs:plot-1}(a). 
    This result complements similar findings of prior work on diffusion transformers~\cite{DBLP:journals/corr/abs-2410-08184}, discrete diffusion models~\cite{DBLP:journals/corr/abs-2512-10858}, and AR SLMs~\cite{cuervo2024scaling}.
    % Thus, we expect a scaling law fit for the validation loss of audio diffusion models.
    % Our observations of expected isoFLOP behavior for the validation loss complement the findings of previous related works where similar investigations are carried out on diffusion transformers~\cite{DBLP:journals/corr/abs-2410-08184} and for discrete diffusion models~\cite{DBLP:journals/corr/abs-2512-10858}. 
    \item \emph{pJSD for $n \in \left\{1,\dots,5\right\}$ shows the expected isoFLOP behavior} at both weak and strong CFG levels.
    For brevity, we only show the isoFLOPs for $1$-gram and $5$-gram pJSD at weak CFG level in~\Cref{fig:subsec:scaling-laws-for-continuous-diffusion-SLMs:IsoFLOP-analysis:000003_isoFLOPs:plot-1}(b, c). 
    Similar behavior was shown for sBLIMP and sStoryCloze metrics for AR SLMs in~\cite{cuervo2024scaling}.
    \ifappleformat
        \input{body/isoflop_summary_figure}
    \fi
    \item \emph{None of the MOS metrics show expected isoFLOP behavior}; see P808-MOS at weak CFG in~\Cref{fig:subsec:scaling-laws-for-continuous-diffusion-SLMs:IsoFLOP-analysis:000003_isoFLOPs:plot-1}(f). 
    \emph{Only two of the four Meta Audiobox Aesthetics components, content enjoyment (CE) and content understanding (CU), show expected isoFLOP behavior, whereas 
    % the other two components %NB(jrp): already clear.
    production complexity (PC) and production quality (PQ) do not}; see~\Cref{fig:subsec:scaling-laws-for-continuous-diffusion-SLMs:IsoFLOP-analysis:000003_isoFLOPs:plot-1}(d, e).
    % and only two Most of the speech perception quality metrics do not show the expected isoFLOP behavior, except for two (content enjoyment (CE) and content understanding (CU))  of the four Meta AudioBox Aesthetics components}.
    % These three components include content enjoyment (CE), content understanding (CU), and production quality (PQ), while the remaining component of production complexity (PC) does NOT demonstrate the expected isoFLOP behavior. 
    % As a representative, we show P808-MOS and the CU component at weak CFG level in~\Cref{fig:subsec:scaling-laws-for-continuous-diffusion-SLMs:IsoFLOP-analysis:000003_isoFLOPs:plot-1} (f, d). 
    % All perceptual quality metrics do not show expected isoFLOP behavior. We observed quick saturation: their values are within the standard deviation of the metric computed for real data (~\Cref{fig:subsec:scaling-laws-for-continuous-diffusion-SLMs:IsoFLOP-analysis:000003_isoFLOPs:plot-1}(e, f)). This is generally positive as models quickly learn to output proper audio quality even under small compute budgets.
    The perceptual quality metrics that do not exhibit expected isoFLOP behavior saturate quickly to within the standard deviation of real-data baselines (\Cref{fig:subsec:scaling-laws-for-continuous-diffusion-SLMs:IsoFLOP-analysis:000003_isoFLOPs:plot-1}(e, f)), suggesting that models efficiently learn to quickly produce reasonable audio quality output with minimal compute.
    % Thus, models learn quickly to output proper audio quality even under small compute budgets.
    % Further, none of the DNSMOS or NISQA metrics demonstrate the expected isoFLOP behavior. 
    % \item We hypothesize that \emph{the lack of observing desired scaling behavior might be caused because the underlying metric might already be saturated}; that is, all trained models have already learned the respective features to a great extent, so much so that the metric values are already very close to their baseline values. 
    % To test this hypothesis, we compute these speech perceptual quality metrics for the ground truth data and plot its mean as the optimal baseline value for the models to achieve, along with a $\pm\sigma$ standard deviation band. 
    % From our isoFLOP investigations, we see that \emph{metrics with baselines that are already saturated do not show the desired scaling behavior, whereas those that are yet to reach their baseline values indeed demonstrate the expected isoFLOP behavior}.
    % As representatives of the former case, isoFLOPs for production complexity and P808-MOS components are shown in~\Cref{fig:subsec:scaling-laws-for-continuous-diffusion-SLMs:IsoFLOP-analysis:000003_isoFLOPs:plot-1} (e, f).
    % Thus, the lack of observing expected isoFLOP behavior in these metrics is not fully unexpected and indicates that our models are learning the underlying features that are gauged by these saturating metrics almost perfectly even at smaller values of compute budgets under consideration.
\end{enumerate}
Given expected isoFLOP behavior for validation loss, pJSD, CE and CU components of Meta Audiobox Aesthetics, we can expect that they all scale with compute and there is a predictive scaling law fit for all of them.

% \begin{figure*}[t!]
%     \centering
%     \includegraphics[width=0.49\textwidth]{plotting/isoflop_curves_vs_D_phoneme_jsd_3gram.png}
%     \hfill
%     \includegraphics[width=0.49\textwidth]{plotting/isoflop_curves_vs_D_eval_loss_min_snr_avg.png}
%     \caption{
%     IsoFLOP curves as a function of dataset size $D$ for (left) phoneme 3-gram
%     Jensen Shannon distance and (right) Min-SNR weighted eval loss.
%     Points are measured runs; dashed curves are the fitted IsoFLOP predictions from
%     Eq.~\eqref{eq:scaling_surface}. Stars denote compute-optimal allocations $(N^{\star}, D^{\star})$
%     implied by the fitted surface at each compute budget. \textbf{NOTE(jrp): figure needs cleaning. font too small. apple-cmap, etc}
%     }
%     \label{fig:isoflop_curves}
% \end{figure*}

%

\subsection{Scaling Law for Validation Loss}
\label{subsec:scaling-laws-for-continuous-diffusion-SLMs:scaling-law-for-validation-loss}

% \noindent For validation loss and other downstream evaluation metrics that demonstrate the required isoFLOP behavior, we set out to fit scaling laws. 
% We start with the discussion of fitting a scaling law for the validation loss.
% For this, we start with the functional form of~\Cref{eq:chinchilla_scaling}, as shown below:
% \begin{equation}
%     \label{eq:scaling-laws-for-continuous-diffusion-SLMs:scaling-law-for-validation-loss:base-scaling-law-eq}
%     L 
%     = 
%     E 
%     +
%     \left(  
%         \frac{A}{N^\alpha} 
%         +
%         \frac{B}{D^\beta}
%     \right)^\gamma
% \end{equation}
% \noindent For each compute budget $C$ and for each model size and dataset size pair $\left(N, D\right)$ that adheres to the compute $C$, we train different models with at least $3$ different seeds.

We start with fitting a scaling law to the validation loss.
For each $\left(N, D\right)$ setting, the average validation loss of all the seeds is used as the representative validation loss.
% of that setting $\left(N, D\right)$. 
To find the optimal parameters $E, A, B, \alpha, \beta, \gamma$ of the scaling law expression of~\Cref{eq:chinchilla_scaling}, we use Huber loss as the learning objective.
For optimization, we use the basin-hopping algorithm~\cite{wales1997global,li1987monte} with the L-BFGS-B method~\cite{byrd1995limited} and 2k iterations. 
The key takeaways~are:

\begin{enumerate}
    % \item \emph{It is necessary to include the overall power $\gamma$ into optimization to be able to achieve a scaling law fit with less than~5\% MRE}. 
    % % Thus, we chose the functional form of $\Cref{eq:scaling-laws-for-continuous-diffusion-SLMs:scaling-law-for-validation-loss:base-scaling-law-eq}$ rather than the vanilla form $L = E + \frac{A}{N^\alpha} + \frac{B}{D^\beta}$ ($\gamma = 1$ being fixed). 
    % The best scaling law fit we find is shown in~\Cref{fig:related_work:header_figure}(a) with coefficients $E=0.0055, A=0.0638, B=29.7667, \alpha=0.3995, \beta=0.5644, \gamma=0.7051$.
    \item \emph{Including the overall power $\gamma$ during optimization is necessary to achieve a scaling law fit with under 5\% MRE.} The best scaling law fit we found, shown in~\Cref{fig:related_work:header_figure}(a), yields the coefficients $E=0.0055, A=0.0638, B=29.7667, \alpha=0.3995, \beta=0.5644, \gamma=0.7051$.
    \item Using these coefficients, we compute several quantitative details of the isoFLOPs: the optimal model size $N^\ast(C)$ and the corresponding optimal dataset size $D^\ast(C)$ needed for any compute budget $C$. 
    As shown in~\Cref{fig:scaling-laws-for-continuous-diffusion-SLMs:scaling-law-for-validation-loss:scaling-laws-quantitative-and-extrapolation}, \emph{the optimal tokens-per-parameter ratio $r^\ast(C)=D^\ast / N^\ast$ decreases with compute budget $C$}. 
    This behavior makes CD SLMs an increasingly efficient option at higher compute levels. This contrasts with AR SLMs using 25Hz SSL tokenization, where $r^\ast(C)$ was shown to increase with compute~\cite{cuervo2024scaling}.\footnote{For lower frame rate tokenization, $r^\ast(C)$ in AR SLMs decreased with compute, behaving similarly to our CD SLMs.}  We note that for any optimal tokens-per-parameter ratio $r^\ast$, there is an \emph{equivalent} text-tokens-per-parameter ratio $r^\ast_{\text{text}}$ which is approximately $r^\ast / 20$ based on our estimations in~\Cref{sec:speech-repres}. Given $r^\ast=245$ at $C=10^{21}$, the equivalent text-tokens-per-parameter ratio is $r^\ast_{\text{text}} \approx 12.25$. This is lower than the compute-optimal ratio for text AR LMs, which is reported to be around $20$~\cite{DBLP:journals/corr/abs-2203-15556} (occurring at $r^\ast = 400$ in our setup). \textit{This indicates that by a compute budget of $10^{21}$ FLOPs, CD SLMs utilize compute more efficiently than both text AR and AR SLMs.}  

    \begin{figure}[t]
        \centering
        \ifappleformat
            \includegraphics[width=0.46\linewidth]{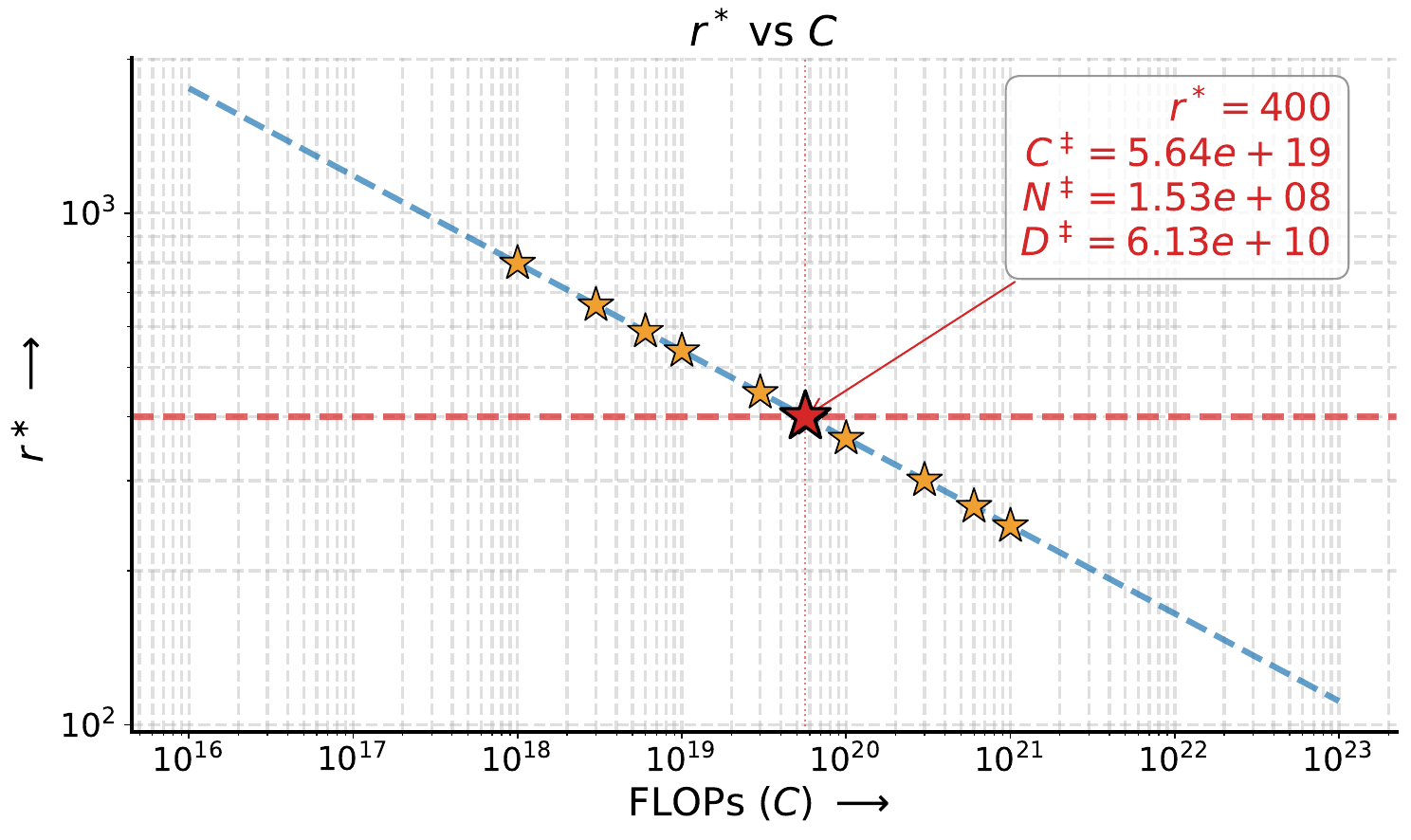}
        \else
            \includegraphics[width=\linewidth]{plotting/000004_scaling_law_fits/eval_loss_min_snr_avg/r_star_vs_C.pdf}
        \fi
        \caption{
            Dependence between optimal tokens-per-parameter ratio $r^\ast = D^\ast/N^\ast$ and compute budget $C$. When $C>C^\ddagger=5.64\cdot 10^{19}$, we observe $r^\ast < 400$.
        }
        \label{fig:scaling-laws-for-continuous-diffusion-SLMs:scaling-law-for-validation-loss:scaling-laws-quantitative-and-extrapolation}
    \end{figure}
    \item \emph{The isoFLOPs tend to get flatter as the compute budget increases}, see~\Cref{fig:related_work:header_figure}(a). 
    We quantify this behavior in two ways.
    Firstly, we compute the curvature of the isoFLOP at its optimum, denoted by $\kappa$, using the scaling law fit. 
    Secondly, we consider for each compute budget $C$ the loss value of $L^\ast + \epsilon$, where $L^\ast$ represents the optimal loss obtained at the compute budget $C$ and $\epsilon$ represents the ``tolerated precision'' in loss. 
    The isoFLOP shape dictates that there will be a range of model sizes, denoted by $\Delta N$, and a range of dataset sizes, denoted by $\Delta D$, such that for any pair $(N^\prime, D^\prime)$ of model size and dataset size in that range (while keeping compute budget fixed to $C$) the validation loss will be in range $L^\prime\in \left[L^\ast, L^\ast + \epsilon\right]$, which we consider to be equivalent from the point of picking the compute optimal $N$ and $D$.
    We set $\epsilon = 1\mathrm{e}{-3}$ for the sake of illustration\footnote{Note that this choice of $\epsilon = 1\mathrm{e}{-3}$ also mimics the precision to which one usually tracks the losses, which helps in getting a better idea of the isoFLOP curvature.}.
    As isoFLOPs get flatter, the $\Delta N$ and $\Delta D$ get larger.
    \Cref{fig:related_work:header_figure}(b) summarizes that \emph{as compute budget increases, isoFLOP curvature at the optimum point decreases over multiple order of magnitudes}. 
    Consequently, \emph{as compute budget increases, the range of model sizes and dataset sizes, where validation loss stays optimal within the predefined tolerant precision $\epsilon$, increases by approximately two orders of magnitude}.
    \textbf{This behavior converts to efficient recipe in practice: at higher computes, we either can use significantly less data or significantly smaller model to achieve loss close enough to the optimum loss.}
\end{enumerate}

\subsection{Scaling Laws for Evaluation Metrics}
\label{subsec:scaling-laws-for-continuous-diffusion-SLMs:scaling-law-for-evaluation-metrics}

\noindent We fit scaling laws for downstream metrics that demonstrate expected isoFLOP behavior using a fused two-stage approach. By plotting the evaluation metrics against validation loss, we realize that the meaningful metrics must saturate in the extremes, i.e. approaching random performance for poorly trained models and optimal values for well-trained ones. Therefore, a sigmoid-like functional form provides a natural mapping from loss to metric.

Confirming our hypothesis, we observe that different metrics exhibit behavior consistent with a sigmoid-like mapping from loss, suggesting that a generalized sigmoid may capture the overall relationship
%
%We observe that these metrics, when plotted against validation loss, follow a generalized sigmoid: saturating at optimal values for low loss and performing random for high loss. We model this mapping using a parametrized sigmoid function as:
%\noindent Fitting scaling laws for downstream evaluation metrics is an interesting but tough challenge on which substantial progress has been made~\cite{DBLP:conf/iclr/Isik00PVK25,DBLP:journals/corr/abs-2512-08894,DBLP:journals/corr/abs-2107-03374,DBLP:journals/corr/abs-2412-04403}.
%As outlined in~\Cref{sec:bck_scaling_laws}, we follow the fused two-stage approach to fitting scaling laws for the downstream evaluation metrics that demonstrate expected isoFLOP behavior. 
%Following previous works, we begin by observing that these metrics of interest when plotted as a function of the validation loss demonstrate a shape that resembles a generalized sigmoid. 
%Note that this is not unexpected; infinitely well-trained models will have a low validation loss and optimal metric values that saturate at the higher limit, whereas totally untrained models will have a high validation loss with metric values at the lower limit.
%Thus, we first assume that a generalized sigmoid with parametrization defined as follows maps the validation loss $L$ to the corresponding metric value $M$:
\begin{equation}
    M 
    = 
    \textrm{sigmoid}\left(L\right)
    =
    \ell 
    +
    \frac{h - \ell}{
        1 + \exp{\left(-k \cdot (L - L_0)\right)}
    }\,,
    \label{eq:scaling-laws-for-continuous-diffusion-SLMs:scaling-law-for-evaluation-metrics:generalized-sigmoid}
\end{equation}
\noindent where $\ell$ and $h$ are the lower and higher metric limits, $L_0$ the sigmoid midpoint, and $k$ the sharpness. Substituting~\Cref{eq:chinchilla_scaling} for $L$ results in the following full downstream scaling law
%\noindent Here, $\ell$ and $h$ represent the lower and higher limits of the metric respectively, $L_0$ represents the half-way point of the sigmoid where the metric reaches the midpoint of its extremes, and $k$ represents the sharpness and directionality factor of the sigmoid. 
%Combining this expression with the functional form of scaling law for validation loss gives us proposal form of scaling laws for the downstream metrics:
\begin{equation}
    M
    =
    \ell 
    +
    \frac{h - \ell}{
        1 + \exp{
            \left(
                -k \cdot \left(
                    E + \left(\frac{A}{N^\alpha} + \frac{B}{D^\beta}\right)^\gamma
                    - 
                    L_0
                \right)
            \right)
        }
    }\,.
    \label{eq:scaling-laws-for-continuous-diffusion-SLMs:scaling-law-for-evaluation-metrics:full-generalized-sigmoid}
\end{equation}
%\noindent In our fused two-stage approach, we use this as the scaling law form for our downstream metrics of interest and jointly optimize for all the coefficients including the \emph{sigmoid coefficients}: $\ell$, $h$, $L_0$, and $k$ as well as \emph{base scaling law parameters}: $E$, $A$, $B$, $\alpha$, $\beta$, $\gamma$. 
%Here, we observe mixed results with interesting observations, which we enumerate below.
All parameters ($\ell, h, L_0, k, E, A, B, \alpha, \beta, \gamma$) are optimized jointly. The key takeaways are following:
\begin{enumerate}
    %\item For $n-$gram pJSD metrics, we observe that the test MRE decreases, and thus, scaling law fits get better as $n$ increases from $1$ to $5$. 
    %As representatives, we demonstrate the scaling law fits for $1-$gram pJSD and $5-$gram pJSD metrics at CFG level of $2$ in~\Cref{fig:subsec:scaling-laws-for-continuous-diffusion-SLMs:scaling-law-for-evaluation-metrics:summary-of-results} (a, b).
    %For instance, the test MRE of the scaling law fit for the $5-$gram pJSD is approximately $1\%$, which is significantly better than that of $1-$gram JSD, which is approximately $4.5\%$.
    %\Cref{fig:subsec:scaling-laws-for-continuous-diffusion-SLMs:scaling-law-for-evaluation-metrics:summary-of-results} (a, b) report the scaling law coefficients corresponding to the best fit that we found with our strategy. 
    %Note that the values of the base scaling law coefficients in these fits are not all exactly the same as those in~\Cref{table:scaling-laws-for-continuous-diffusion-SLMs:scaling-law-for-validation-loss:scaling-laws-quantitative-and-extrapolation} (a).
    %This indicates that one or more assumptions in our fused two-step approach are suboptimal or incorrect, and that there is a scope for finding better scaling law fits in terms of better optimization, or better fit for mapping validation loss to the metric value, or a better direct approach.

    \item \emph{For $n$-gram pJSD, scaling law fits improve with increasing $n$}: the $5$-gram fit achieves ${\sim}1\%$ test MRE versus ${\sim}4.5\%$ for $1$-gram, see \Cref{fig:subsec:scaling-laws-for-continuous-diffusion-SLMs:scaling-law-for-evaluation-metrics:summary-of-results}(a,b). This is intuitive, as higher-order $n$-grams capture more structured phonotactic patterns that correlate more tightly with the training loss.
    
    Moreover, we observe that the base scaling law coefficients obtained from these downstream fits do not exactly match those of the validation loss fit, indicating that either the sigmoid mapping or the joint optimization introduces bias. Improving the functional form, the optimization strategy, or exploring direct fitting approaches remain promising directions.

    \input{body/eval_scaling_law}
    \item \emph{Content enjoyment (CE) and content understanding (CU) components of Meta Audiobox Aesthetics also exhibit scaling laws with low MRE fits.} \Cref{fig:subsec:scaling-laws-for-continuous-diffusion-SLMs:scaling-law-for-evaluation-metrics:summary-of-results}(c) shows the representative fit for CU at weak CFG level.
    
    %\item We also observe that we can get low MRE good scaling law fits for the three admissible components of MetaAudioBoxAesthetics metric, albeit with the same issue of significantly different base scaling law parameters. 
    %A representative scaling law fit for the content understanding component at CFG level $2$ is shown in~\Cref{fig:subsec:scaling-laws-for-continuous-diffusion-SLMs:scaling-law-for-evaluation-metrics:summary-of-results} (c). 

    \item Since Meta Audiobox Aesthetics components have real-data baselines including mean and standard deviation of metric values on real speech, we can extrapolate the optimal metric value as a function of compute and assess whether models can approach real-data quality, see \Cref{fig:subsec:scaling-laws-for-continuous-diffusion-SLMs:scaling-law-for-evaluation-metrics:summary-of-results}(d). We find that \emph{optimal values saturate and do not reach the $\pm\sigma$ baseline region}, suggesting that certain features required to match real-data quality may not be learnable by our CD SLMs regardless of compute budget (we may also loose some model quality due to reconstruction error of the vocoder).

    This conclusion raises two caveats: \emph{(i)} it assumes the functional form is correct and the optimization is sufficient -- experimenting at wider compute ranges or with alternative functional forms may result in different conclusions; \emph{(ii)} if it does hold, it implies that CD SLMs have inherent representational limitations, potentially necessitating stronger inductive biases, richer data representations, or text-based conditioning to bridge the gap.

\end{enumerate}

\FloatBarrier

\providecommand{\ablationinput}{body/03_06_ablations_v3}

% NB(jrp): cut for space if needed
% \def\ablationinput{body/03_04_ablations}  % New within 8 pages [all]
% \def\ablationinput{body/03_05_ablations_v2}  % Ultra compact.
% \def\ablationinput{body/03_06_ablations_v3}  % More compact
\input{\ablationinput}

\FloatBarrier

% \section{Scaling of Continuous Diffusion SLM to a 16B Parameter Model}
\section{Scaling Continuous Diffusion SLMs to 16B Parameters}
\label{sec:aux_cond}
\begin{figure}[htbp]
  \centering
  \includegraphics[width=0.7\linewidth]{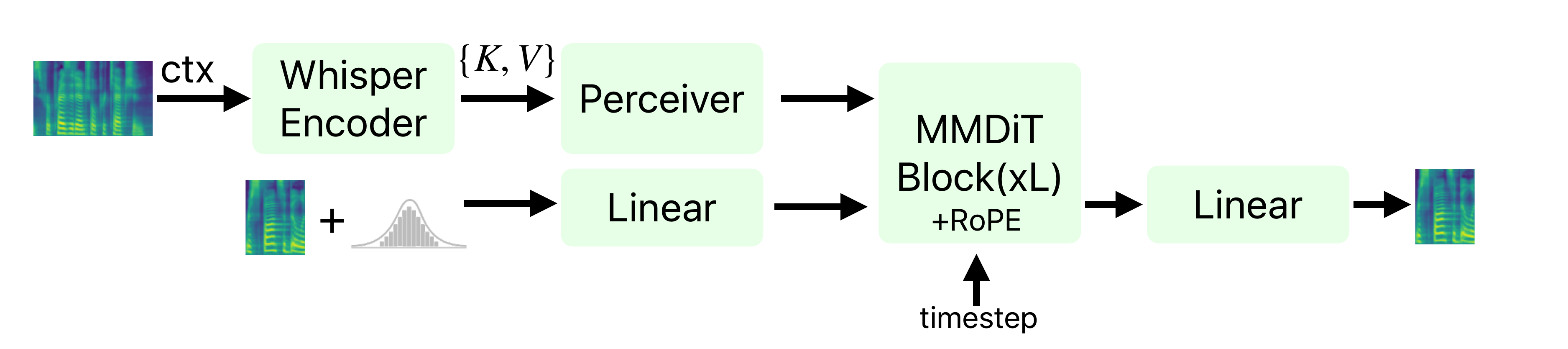}
  % \caption{Continuous diffusion SLM overview where model is based on log mel-filterbanks speech representations and \gls{mmdit} architecture, but it is conditioned on Whisper representations further downsampled by Perceiver.}
  \caption{Whisper conditioned CD SLM architecture.}
  \label{fig:aux_model}
\end{figure}
%
% NB(jrp): rewrite + tightened.
\noindent The scaling law established in \Cref{sec:scaling-laws-for-continuous-diffusion-SLMs} estimates an irreducible loss $E$ (\Cref{eq:chinchilla_scaling}), which is the asymptotic minimum as $N$ and $D$ approach infinity. Because scaling coefficients depend on architecture and data representation \cite{kaplan2020scaling}, our base \gls{mmdit} model's finite-context log-mel filterbanks impose a structural lower bound on performance. Recent findings demonstrate that richer, superposition-exhibiting data representations yield sharper, more robust scaling \cite{liu2025superposition}. Therefore, we hypothesize that information dense conditioning can improve the scaling trajectory and lower this bound empirically.
% theoretical floor.

We introduce a modified architecture (\Cref{fig:aux_model}) integrating auxiliary conditioning from a frozen pretrained Whisper-large-v3 encoder \cite{radford2023robust} to provide a higher level speech context. While Whisper is trained on speech-text pairs, we use it strictly as a frozen feature extractor to assess whether richer representations improve scaling, independent of their training origin. To manage the expanded input of 300s of context generating a 60s continuation, we employ a Perceiver \cite{DBLP:conf/iclr/JaegleBADIDKZBS22} for learned temporal downsampling to a deterministic 4096 tokens. Scaling this architecture to 16B parameters, we train on tens of millions of hours of unfiltered conversational speech from SpeechCrawl.
% (SpeechCrawl audio $>$6 minutes).

Crucially, this model achieves a validation loss below the irreducible loss $E$ estimated for our base architecture, as summarized in \Cref{tab:aux_model}. 
\input{body/16b_table}
This confirms that the lower bound is representation and model dependent rather than a fundamental limit of the data distribution. While the model produces emotive, prosodic, and multilingual speech with improved lexical word $n$-grams (see supplementary material for several examples of generated speech), long-form linguistic coherence remains elusive.
\textit{These findings indicate that advancing SLMs requires a systematic exploration of novel architectures, data representations, and conditioning strategies to catalyze the emergence of linguistic structure.}

\FloatBarrier

%% file: body/isoflop_summary_figure.tex
\widetemplatefigurebegin
    \centering
    % Validation Loss.
    \begin{subfigure}[b]{0.32\textwidth}
        \centering
        \includegraphics[width=\linewidth]{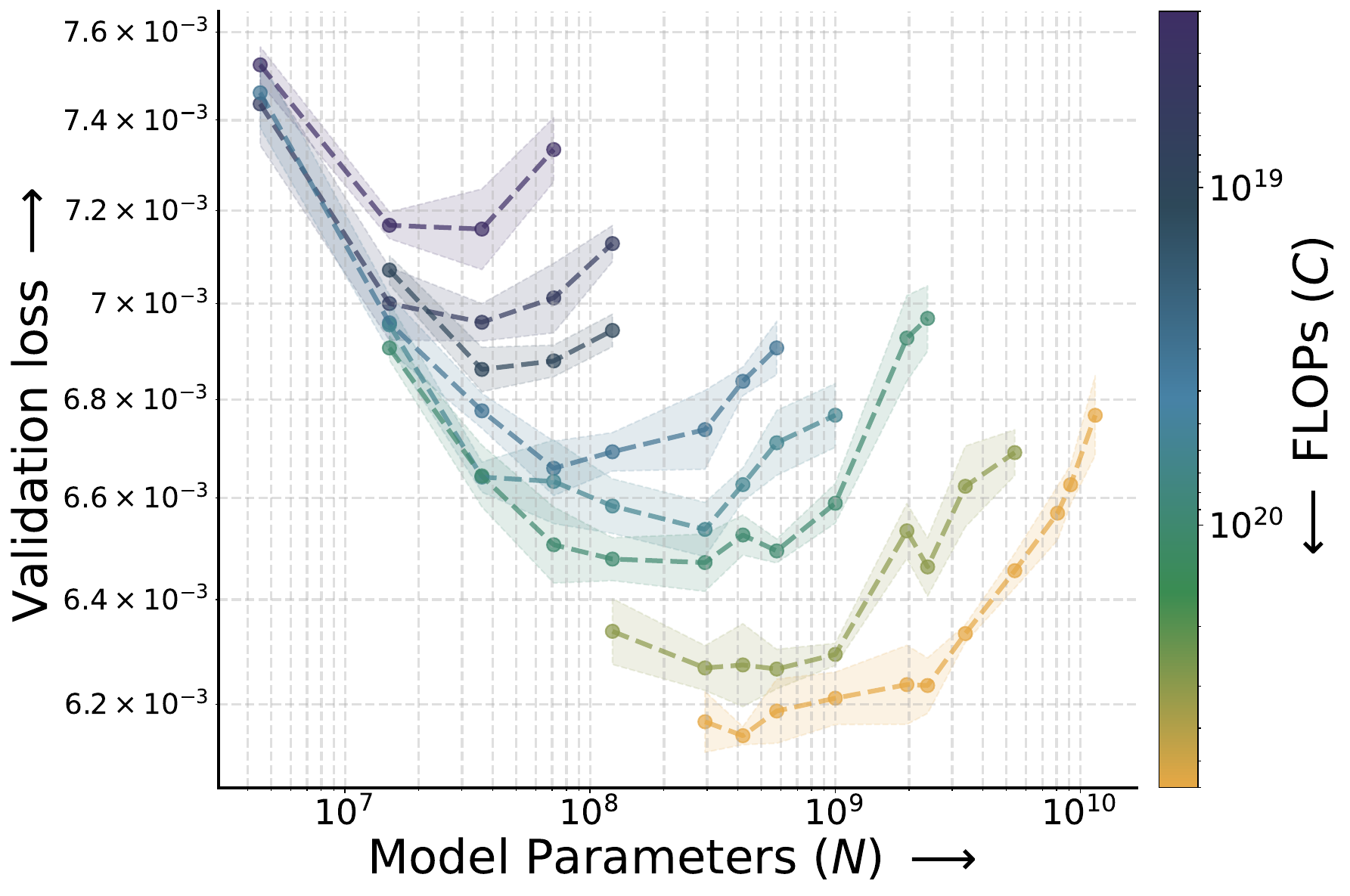}
        \caption*{(a) Validation loss.}
        \label{fig:scaling-laws-for-continuous-diffusion-SLMs:IsoFLOP-analysis:000003_isoFLOPs:eval_loss_min_snr_avg:A}
    \end{subfigure}
    \hfill
    % 1-gram Phoneme JSD.
    \begin{subfigure}[b]{0.32\textwidth}
        \centering
        \includegraphics[width=\linewidth]{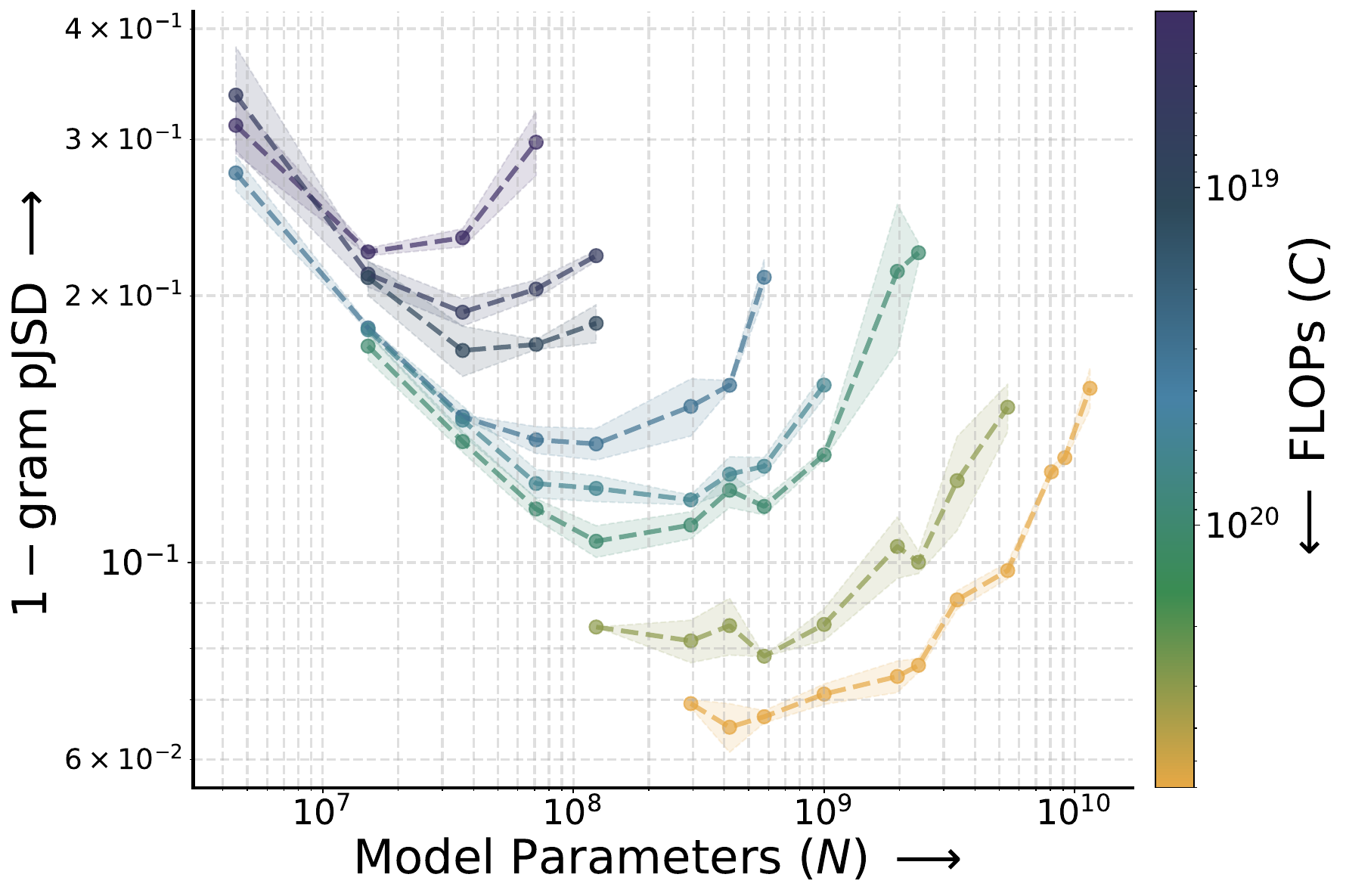}
        \caption*{(b) $1-$gram pJSD.}
        \label{fig:scaling-laws-for-continuous-diffusion-SLMs:IsoFLOP-analysis:000003_isoFLOPs:PhonemeCoverageDivergence_1gram_JSD:B}
    \end{subfigure}
    \hfill
    % 5-gram Phoneme JSD.
    \begin{subfigure}[b]{0.32\textwidth}
        \centering
        \includegraphics[width=\linewidth]{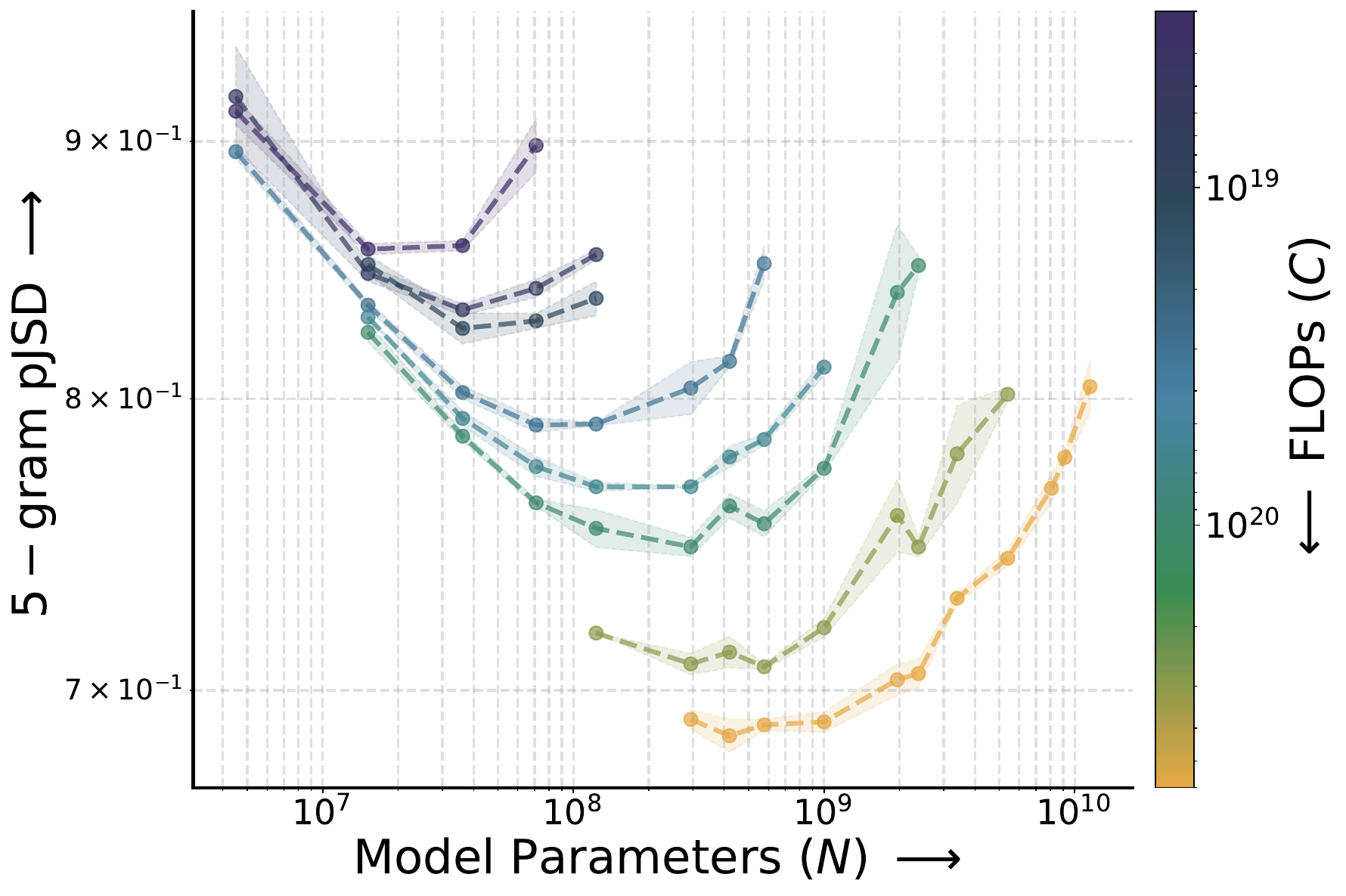}
        \caption*{(c) $5$-gram pJSD.}
        \label{fig:scaling-laws-for-continuous-diffusion-SLMs:IsoFLOP-analysis:000003_isoFLOPs:PhonemeCoverageDivergence_5gram_JSD:C}
    \end{subfigure}
    \\
    % MetaAudioBoxAesthetics Content Understanding (CU).
    \begin{subfigure}[b]{0.32\textwidth}
        \centering
        \includegraphics[width=\linewidth]{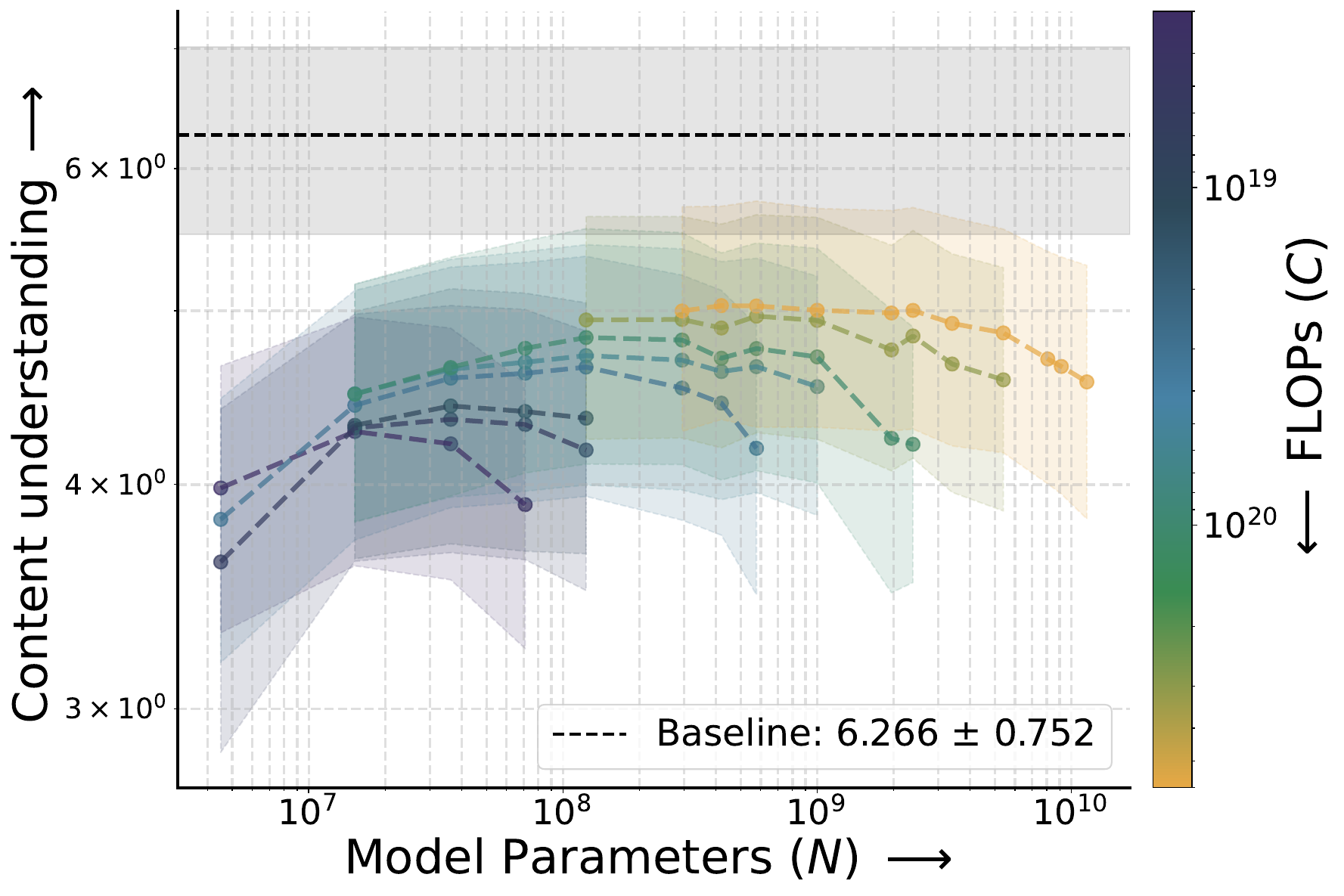}
        \caption*{(d) Content Understanding.}
        \label{fig:scaling-laws-for-continuous-diffusion-SLMs:IsoFLOP-analysis:000003_isoFLOPs:MetaAudioBoxAesthetics_CU:D}
    \end{subfigure}
    \hfill
    % MetaAudioBoxAesthetics Production Complexity (PC).
    \begin{subfigure}[b]{0.32\textwidth}
        \centering
        \includegraphics[width=\linewidth]{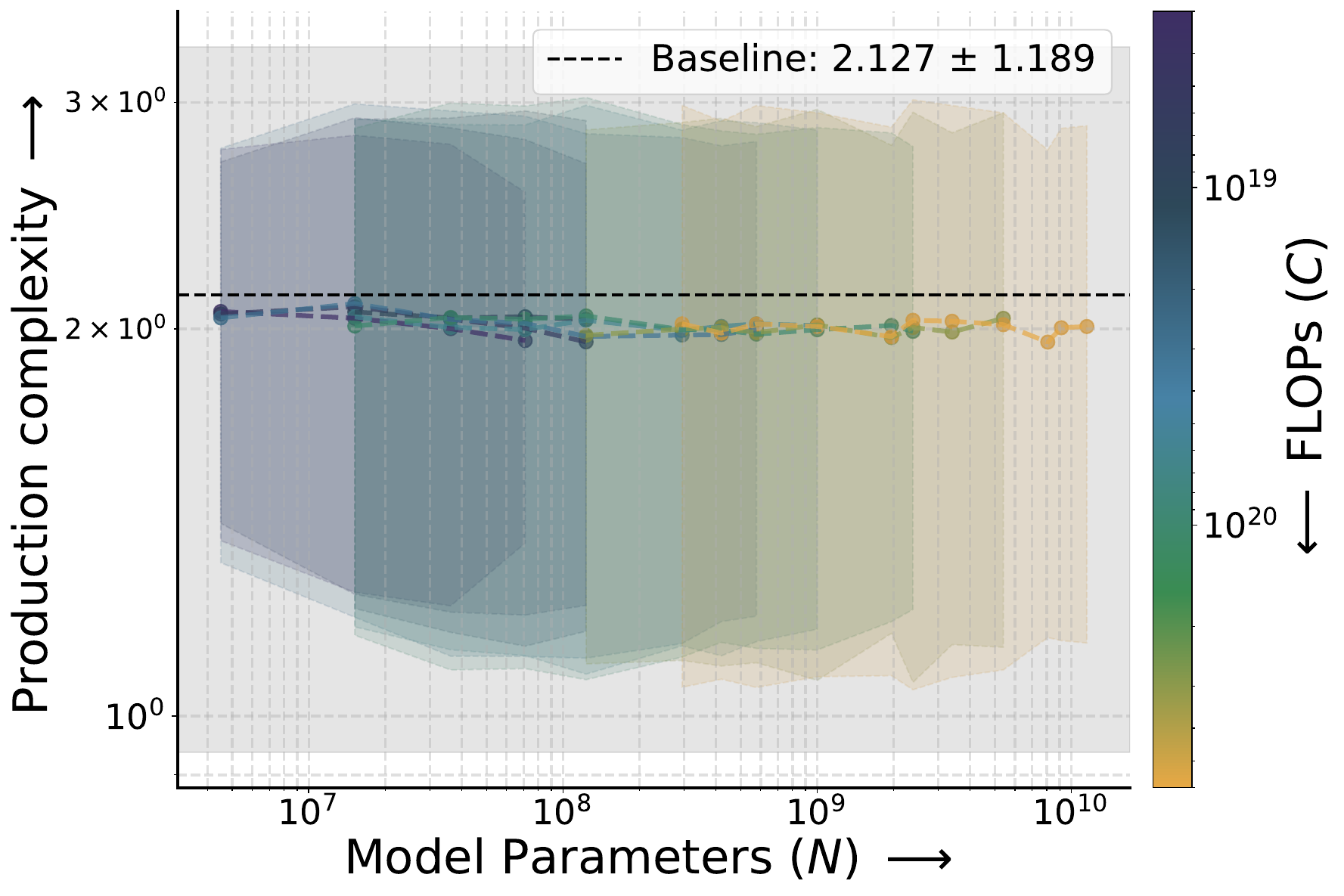}
        \caption*{(e) Production Complexity.}
        \label{fig:scaling-laws-for-continuous-diffusion-SLMs:IsoFLOP-analysis:000003_isoFLOPs:MetaAudioBoxAesthetics_PC:E}
    \end{subfigure}
    \hfill
    \begin{subfigure}[b]{0.32\textwidth}
        \centering
        \includegraphics[width=\linewidth]{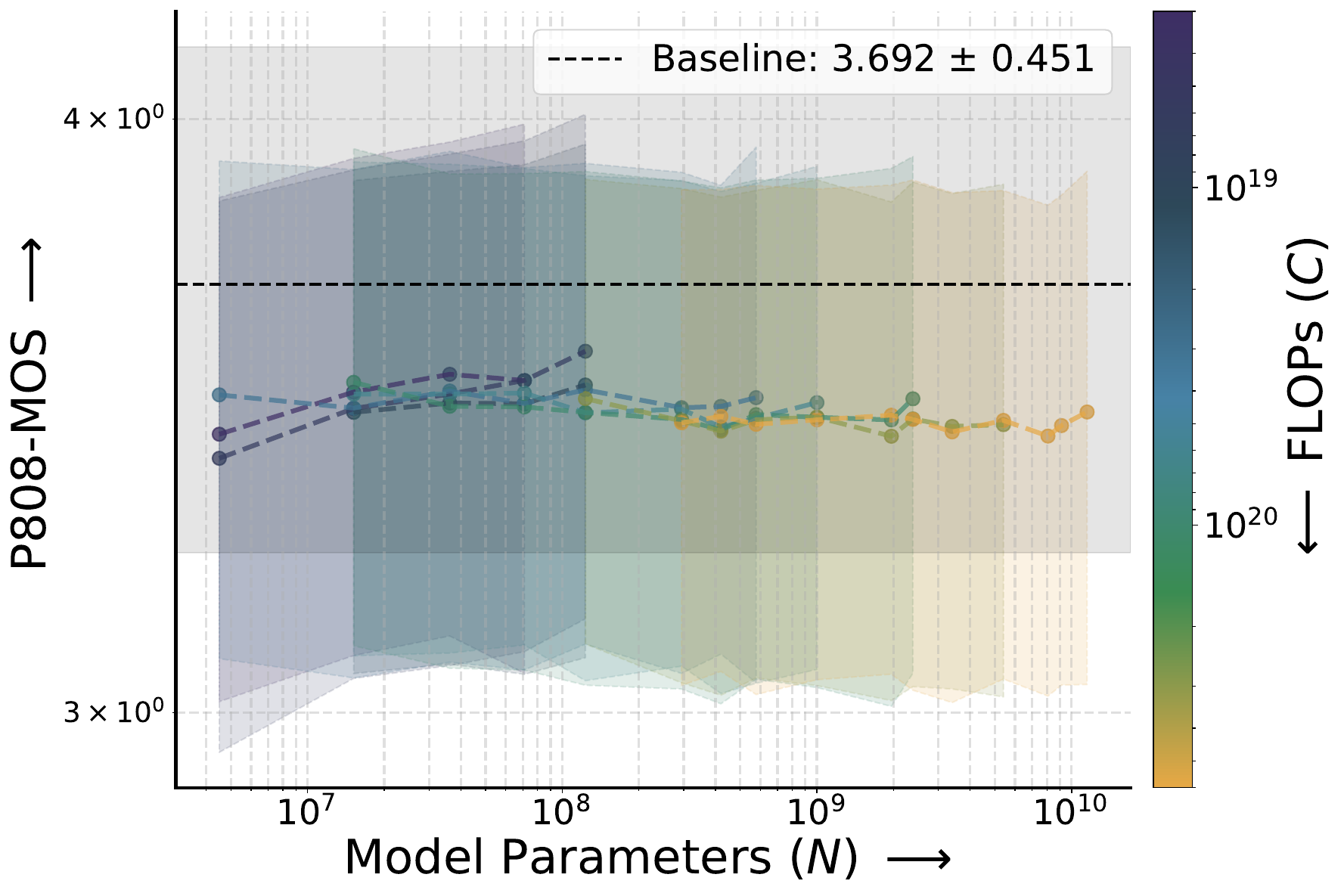}
        \caption*{(f) P808-MOS.}
        \label{fig:scaling-laws-for-continuous-diffusion-SLMs:IsoFLOP-analysis:000003_isoFLOPs:DNSMOS_P808_MOS:F}
    \end{subfigure}
    \caption{
        IsoFLOP curves at weak CFG level. (a-c) Validation loss, 1-gram pJSD, and 5-gram pJSD (lower-is-better) exhibit expected isoFLOP scaling, a trend consistent across all $n$-grams and CFG levels.
        (d) The content understanding (CU) component of Meta Audiobox Aesthetics (higher-is-better) also scales predictably, alongside content enjoyment.
        (e-f) In contrast, production quality and production complexity components of Meta Audiobox Aesthetics, alongside all automatic MOS do not show expected scaling. Instead, they quickly saturate within the $\pm\sigma$ range of the real-data baseline (indicated by the black line and gray fill).
    }
    \label{fig:subsec:scaling-laws-for-continuous-diffusion-SLMs:IsoFLOP-analysis:000003_isoFLOPs:plot-1}
\widetemplatefigureend

%% file: body/eval_scaling_law.tex
\widetemplatefigurebegin
    \centering
    % 1-gram Phoneme JSD.
    \begin{subfigure}[b]{\ifappleformat0.472\linewidth\else0.430\linewidth\fi}
        \centering
        \includegraphics[width=\linewidth]{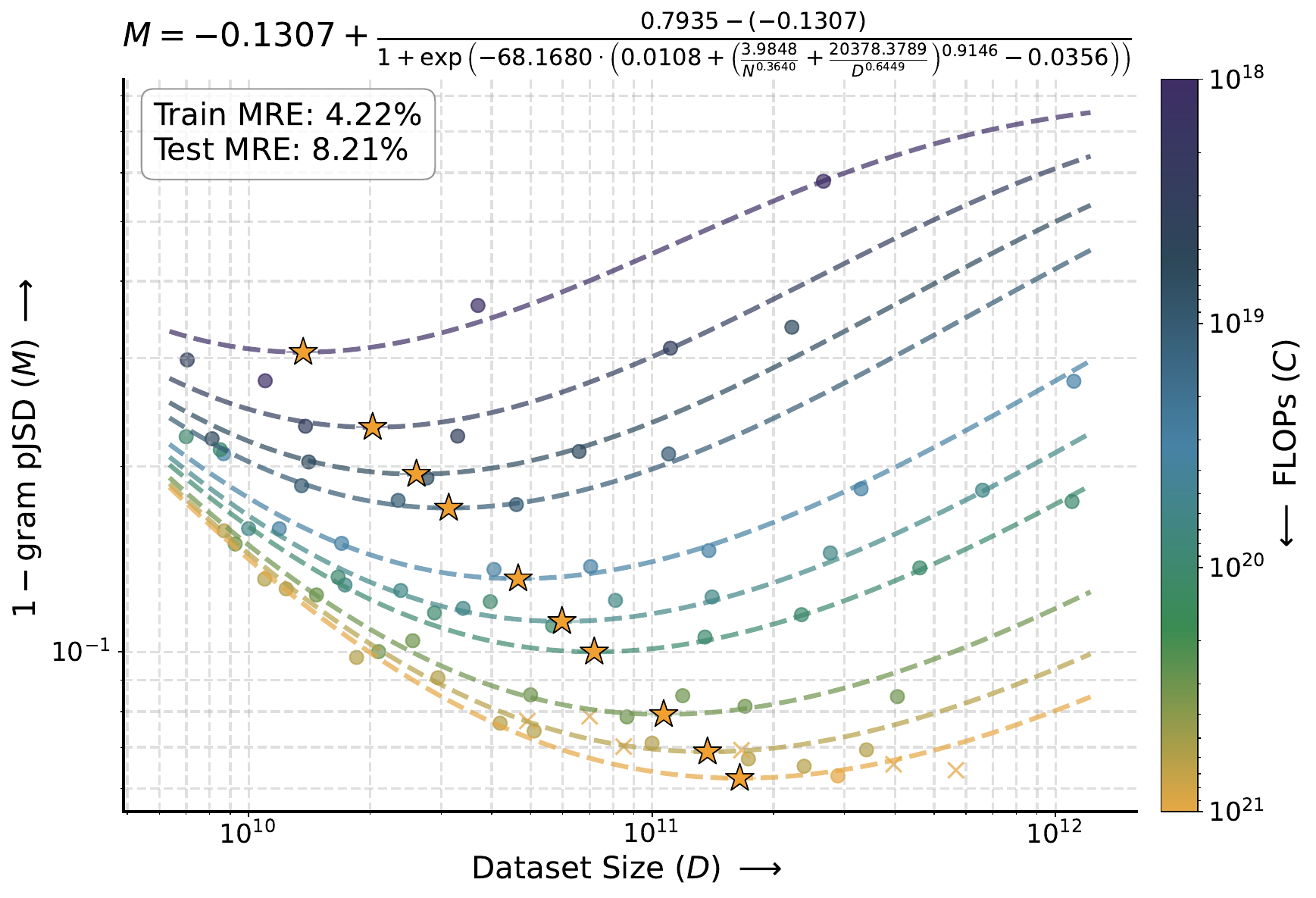}    
        \caption*{(a)}    
        \label{fig:subsec:scaling-laws-for-continuous-diffusion-SLMs:scaling-law-for-evaluation-metrics:summary-of-results:A}
    \end{subfigure}
    \hspace{\ifappleformat0.004\linewidth\else0.008\linewidth\fi}
    % 5-gram Phoneme JSD.
    \begin{subfigure}[b]{\ifappleformat0.472\linewidth\else0.430\linewidth\fi}
        \centering
        \includegraphics[width=\linewidth]{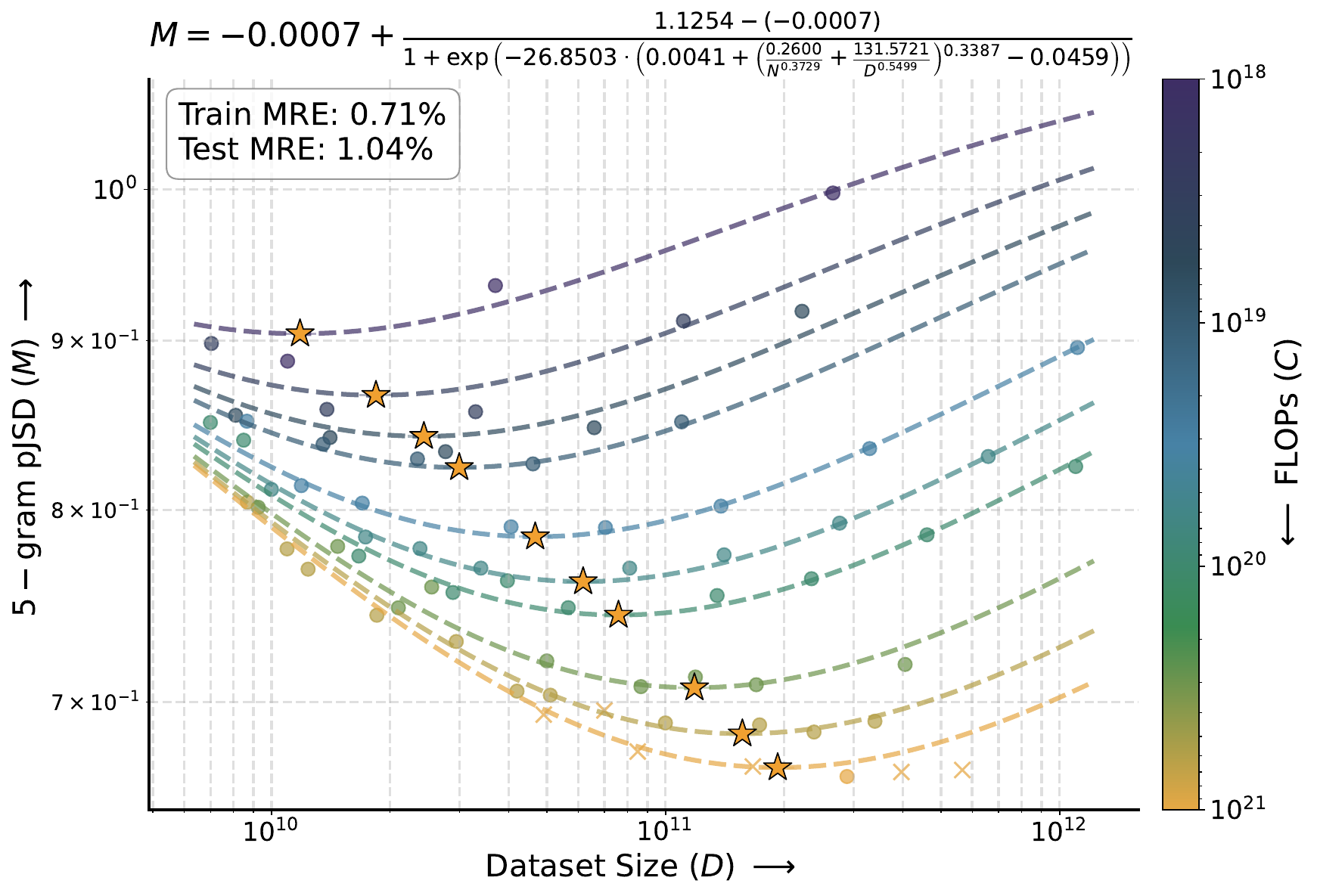}
        \caption*{(b)}     
        \label{fig:subsec:scaling-laws-for-continuous-diffusion-SLMs:scaling-law-for-evaluation-metrics:summary-of-results:B}
    \end{subfigure}
    \par\ifappleformat\smallskip\fi
    % MetaAudioBoxAesthetics Content Understanding (CU).
    \begin{subfigure}[b]{\ifappleformat0.472\linewidth\else0.430\linewidth\fi}
        \centering
        \includegraphics[width=\linewidth]{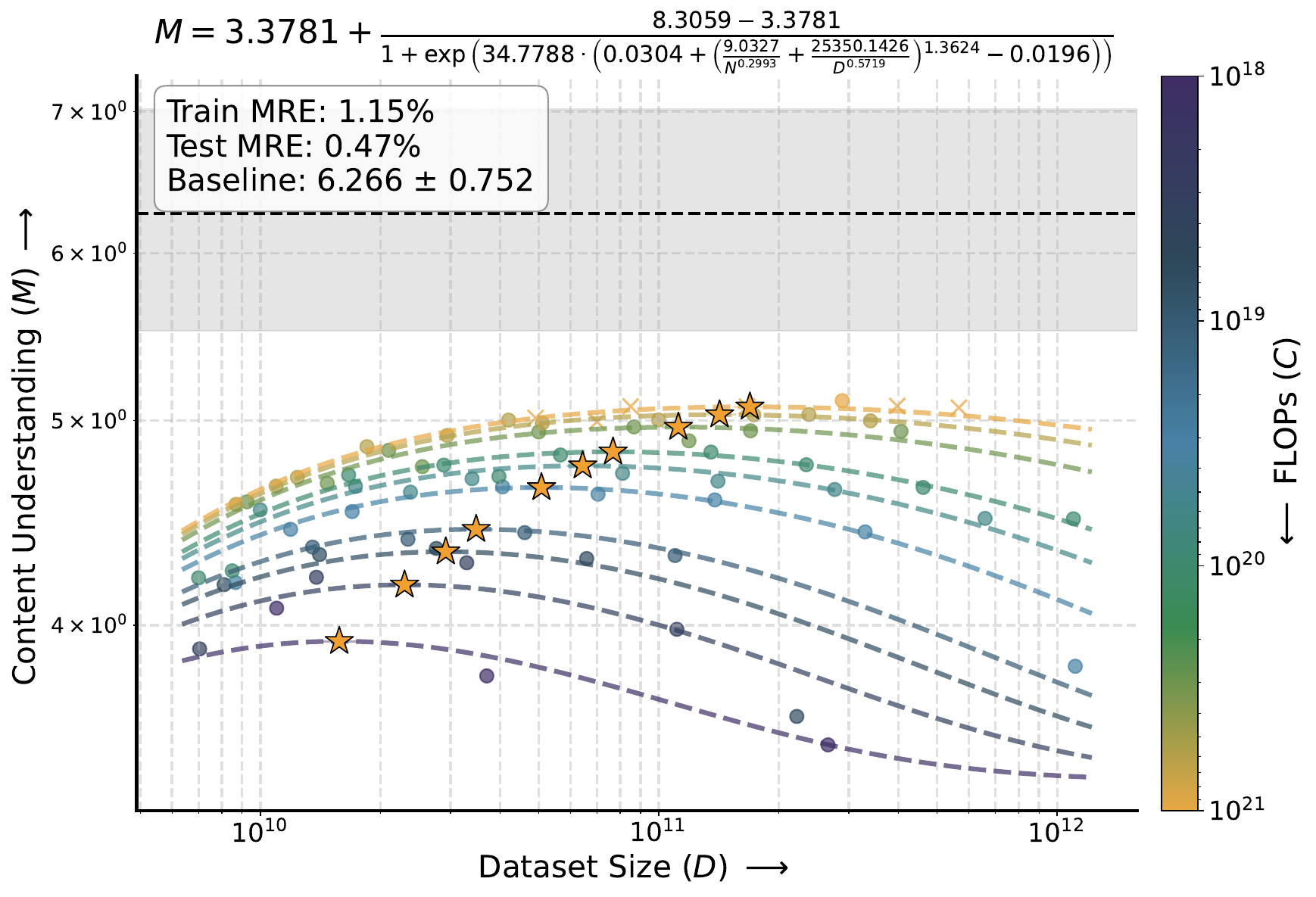}
        \caption*{(c)}
        \label{fig:subsec:scaling-laws-for-continuous-diffusion-SLMs:scaling-law-for-evaluation-metrics:summary-of-results:C}
    \end{subfigure}
    \hspace{\ifappleformat0.004\linewidth\else0.008\linewidth\fi}
    % MetaAudioBoxAesthetics Production Complexity (PC).
    \begin{subfigure}[b]{\ifappleformat0.472\linewidth\else0.430\linewidth\fi}
        \centering
        \includegraphics[width=\linewidth]{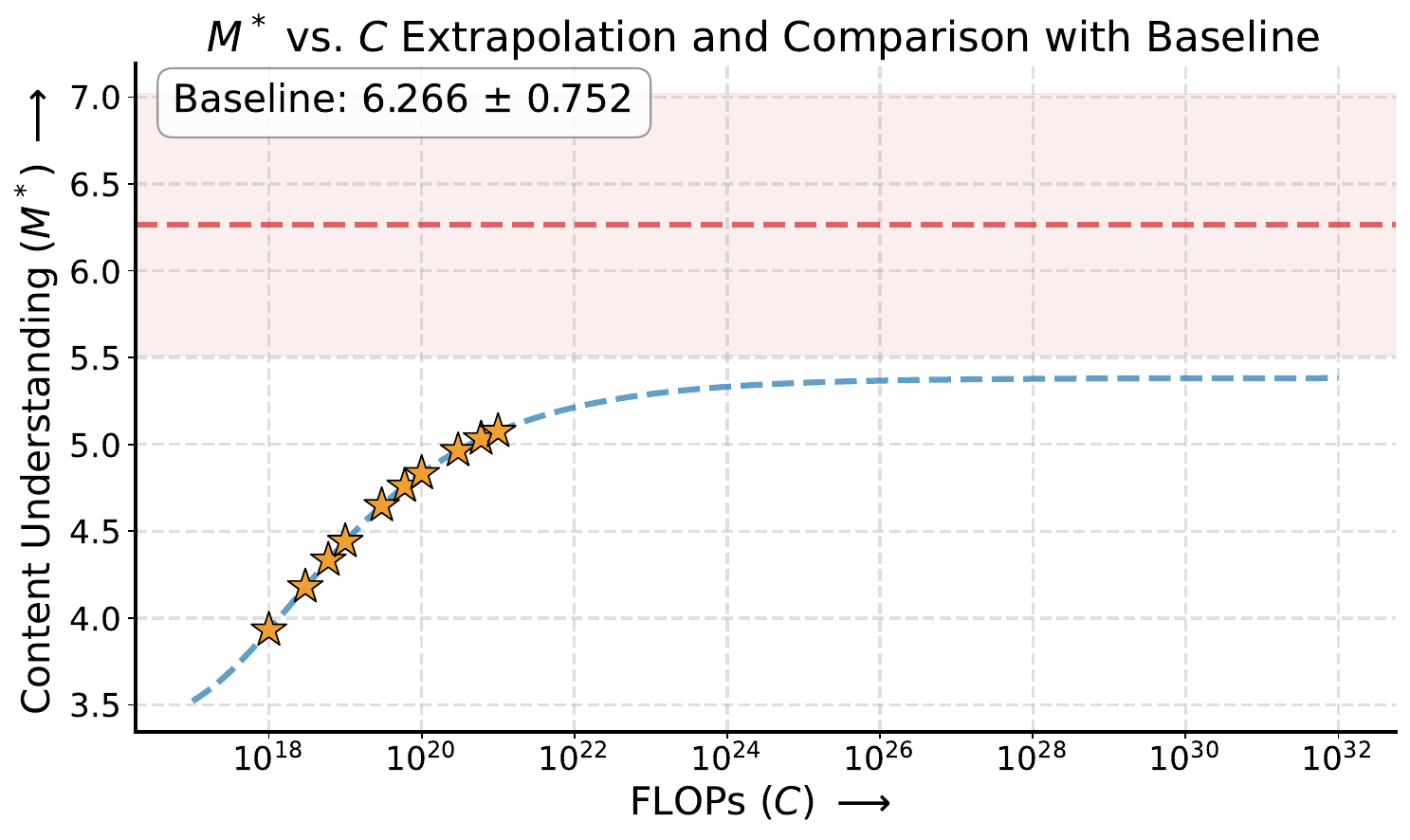}
        \caption*{(d)}     
        \label{fig:subsec:scaling-laws-for-continuous-diffusion-SLMs:scaling-law-for-evaluation-metrics:summary-of-results:D}
    \end{subfigure}
    % \caption{
    %     (a, b) The fused two-stage scaling law fit for $1-$gram pJSD and $5-$gram pJSD metrics respectively. 
    %     We observe that higher $n-$grams show a better scaling law fit with lower MRE.
    %     We show $1-$gram and $5-$gram fits as representatives.
    %     (c) Analogously, we attempt fused two-stage scaling law fitting for admissible components of MetaAudioBoxAesthetics metric. 
    %     As a representative, we show the scaling law fit for the content understanding metric component.
    %     (d) The plot of the optimal content understanding value $M^\ast$ as a function of the compute budget $C^\ast$. 
    %     Assuming that the scaling law fitting is correct and would continue to hold with the same form, we see that no matter what compute budget is used, the models may not approach metric component values close to their baselines (within $\pm\sigma$ region of the baseline).
    %     As a representative, we show the example plot of the optimal content understanding metric value as a function of compute budget at CFG level $2$. 
    %     However, we see exactly the same trend with other admissible MetaAudioBoxAesthetics metric components.
    % }
    \caption{
        (a, b) Fused two-stage scaling law fits for $1$-gram and $5$-gram pJSD metrics. Higher $n$-grams consistently yield better fits with lower mean relative error (MRE).
        (c) Analogous scaling law fit for the content understanding component of the Meta Audiobox Aesthetics metric.
        (d) Extrapolated optimal content understanding $M^\ast$ versus compute budget $C^\ast$ at weak CFG level. Assuming the functional form holds, models may not reach real data quality (within the $\pm\sigma$ region) strictly through compute scaling. This saturation trend remains consistent across all other admissible Meta Audiobox Aesthetics components.
    }
    \label{fig:subsec:scaling-laws-for-continuous-diffusion-SLMs:scaling-law-for-evaluation-metrics:summary-of-results}
\widetemplatefigureend

%% file: body/03_06_ablations_v3.tex
\section{Continuous Diffusion SLM Ablations}
\label{sec:ablation_summary}
\ifappleformat
  \input{body/ablation_summary_figure}
\fi

To understand the sensitivity of our CD SLM to different design choices, we conduct a systematic ablation study across four axes:

\begin{itemize}
    \item \textit{Training duration} measures the cumulative hours of audio the model is trained on. It spans from 0.25M to 1.5M hours in increments of 0.25M hours.
    \item \textit{Temporal patch size $k$} (analogous to spatial patching in vision transformers) folds the temporal dimension by a factor of $k$ while proportionally expanding the channel dimension. This reduces sequence length and computational cost but potentially sacrifices fine-grained temporal resolution. Patch sizes span from 1 to 6 in increments of 1.
    \item \textit{Noise schedule} determines how the signal-to-noise ratio (SNR) changes over diffusion timesteps, and thus which noise levels dominate the learning problem and the denoising trajectory at sampling time. We evaluate three schedules (linear, cosine, exponential), each with and without zero terminal SNR (enforcing complete signal destruction at $t=T$)~\cite{DBLP:conf/wacv/LinLLY24}. 
    \item \textit{Number of diffusion timesteps $T$} determines the granularity of the noise level discretization during training. Finer discretization (larger $T$) provides more precise noise level targets but increases the complexity of the learning problem. We train multiple models with $T \in \{100, 500, 1000, 2000, 4000\}$ and evaluate them using $100$ steps at generation time.
\end{itemize}

Each ablation isolates a single variable while holding others fixed at default values. 
To assess the interaction with inference-time guidance, we report results for both weak and strong CFG scales.
For all ablations, we train a model with $d_{\text{emb}}=1024$ and 8 layers for 512,000 hrs using 100 inference NFE steps.
% ({\color{red} XXX parameters + XXX hyper parameters on training hours, schedule, patch and diffusion steps defaults}). 
We tune the base model hyperparameters using a sweep over learning rate, weight decay, and Adam parameters $\beta_1, \beta_2$, and $\epsilon$.
\ifappleformat\else
  \input{body/ablation_summary_figure}
\fi
Figure~\ref{fig:ablation_cross} synthesizes results across all ablation studies, plotting the distribution of each evaluation metric by ablation type.

\textbf{General Takeaways}~~~
Figure~\ref{fig:ablation_cross} shows that the choice of noise schedule has the largest impact on perceptual quality. This is expected, given that perceptual quality measures signal fidelity, while the noise schedule directly dictates the noise levels in the audio.
Conversely, training duration exhibits the largest impact on languageness, as well as on the content enjoyment and understanding metrics of Meta Audiobox Aesthetics. The latter is consistent with the results in~\Cref{sec:scaling-laws-for-continuous-diffusion-SLMs} on scaling laws.

\textbf{Patch Size Takeaways}~~~
A consistent pattern emerges: as patch size increases (reducing temporal resolution), all metrics degrade. This demonstrates that temporal resolution is critical for high-fidelity and intelligent audio generation. While larger patch sizes offer computational savings, the resulting quality degradation may be unacceptable for applications requiring natural prosody and fine temporal detail. 

\textbf{Noise Schedule Takeaways}~~~
First, the cosine schedule is consistently uncompetitive, trailing linear and exponential alternatives on perceptual quality metrics. 
Second, zero terminal SNR is most beneficial when combined with the linear schedule, suggesting that explicitly training for complete signal destruction improves robustness at the high-noise end of the trajectory.

%% file: body/ablation_summary_figure.tex
\ifappleformat
  \begin{center}
    \captionsetup{type=figure}
    \centering
    \includegraphics[width=0.80\linewidth]{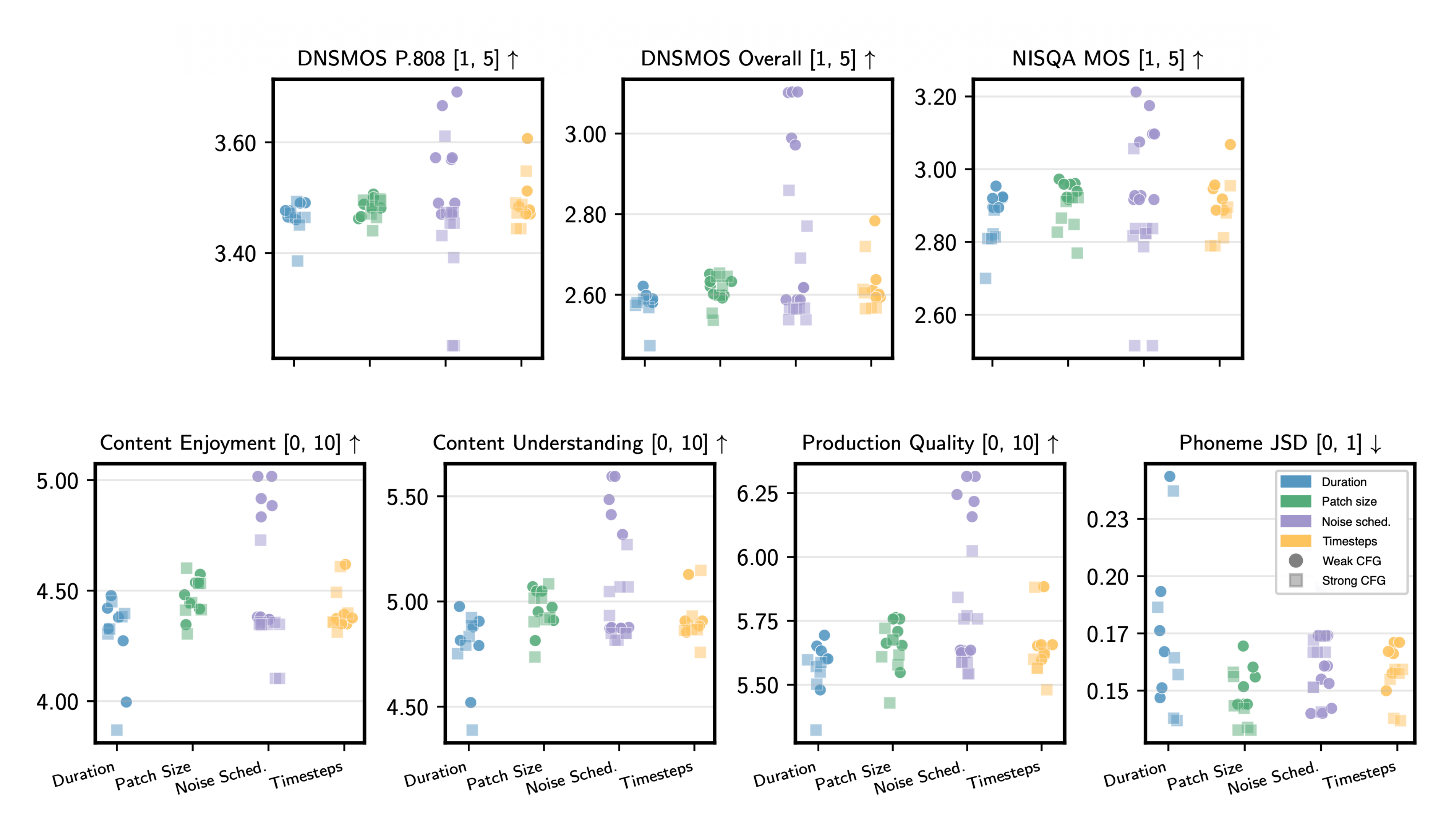}
    \captionof{figure}{Cross-ablation comparison showing metric distributions across all studies. Noise schedule choice exhibits the largest impact on perceptual quality, while the duration exhibits the largest impact on languageness.
    % and timestep ablations show tighter distributions indicating greater robustness to suboptimal settings. 
    % {\color{red} TODO: remove FAD; fix legend to weak/strong CFG}
    }
    \label{fig:ablation_cross}
  \end{center}
\else
  \begin{figure}[t]
    \centering
    \includegraphics[width=\linewidth]{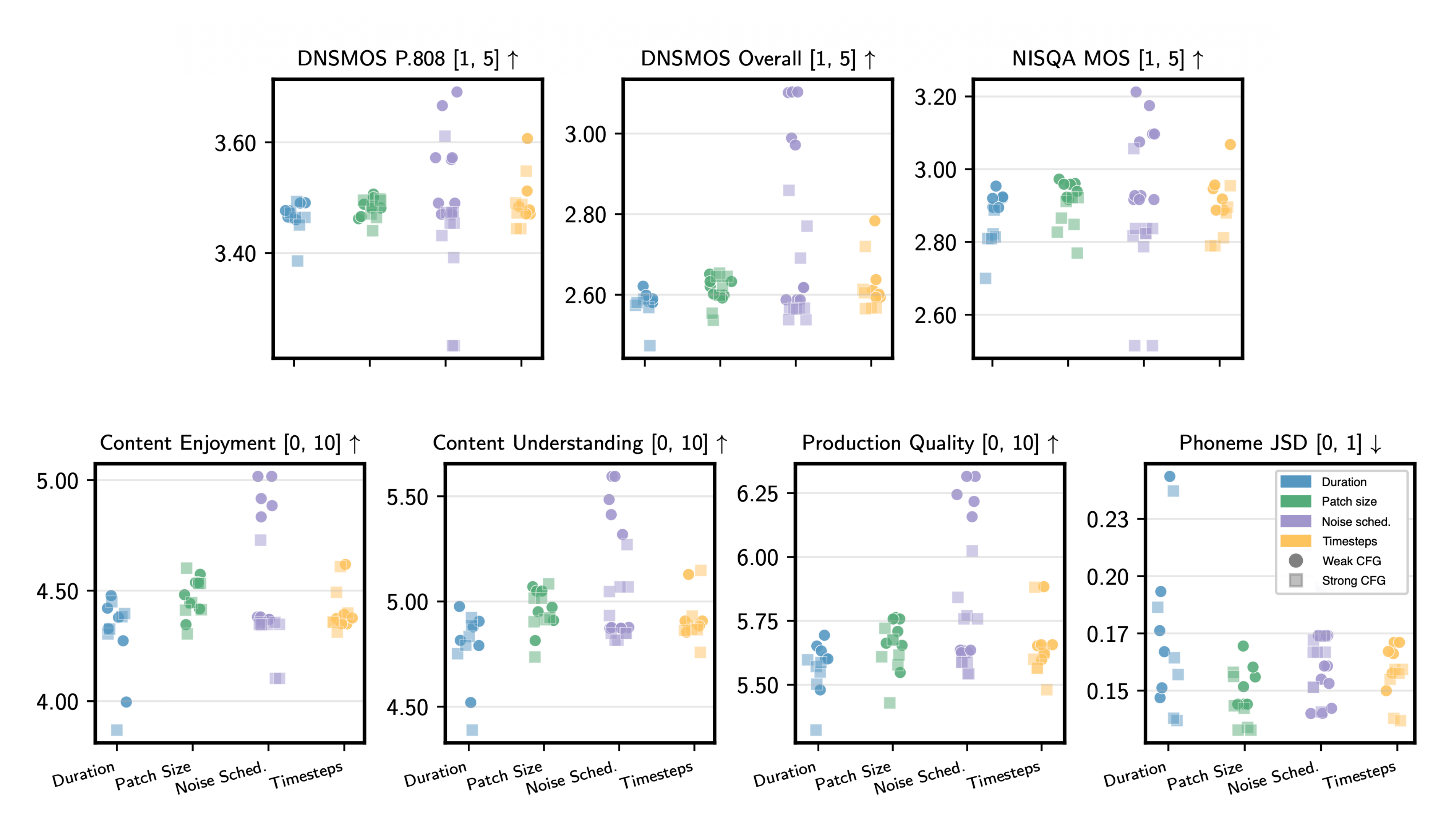}
    \caption{Cross-ablation comparison showing metric distributions across all studies. Noise schedule choice exhibits the largest impact on perceptual quality, while the duration exhibits the largest impact on languageness.}
    \label{fig:ablation_cross}
  \end{figure}
\fi

%% file: body/16b_table.tex
\ifappleformat
\begin{table}[!htbp]
\else
\begin{table}[!t]
\fi
\centering
\footnotesize
\setlength{\tabcolsep}{5pt}
\caption{16B CD SLM vs. best run from the scaling law trials. %{\color{red} todo - fix names of metrics + transpose table to save space}}
}
\begin{tabular}{l | c | c | c | c | c | c}
\toprule
    {}
    & 
    \shortstack{$\textrm{C}=10^{21}$\\$\textrm{CFG}=2$}
    &
    \shortstack{$\textrm{C}=10^{21}$\\$\textrm{CFG}=4$}
    & 
    \shortstack{$16\textrm{B}$\\$\textrm{CFG}=2$}
    & 
    \shortstack{$16\textrm{B}$\\$\textrm{CFG}=4$}
    % & 
    % \shortstack{$17\textrm{B}$\\$\textrm{cfg}=2$}
    % & 
    % \shortstack{$17\textrm{B}$\\$\textrm{cfg}=4$}
    \\
\midrule
    loss
    & 
    0.0061
    & 
    0.0061
    & 
    0.0047
    &     
    0.0047
    % &
    % 0.0048
    % &
    % 0.0048
\\
    CE
    & 
    {4.5767}
    & 
    {4.5545}
    &
    {4.7207}
    &
    {4.7712}
    % &
    % {4.4650}
    % &
    % {4.7712}
\\
    CU
    & 
    {5.1093}
    & 
    {5.0746}
    &
    {5.4809}
    &
    {5.2965}
    % &
    % {5.1787}
    % &
    % {5.2965}
\\
    PQ
    & 
    {5.6893}
    & 
    {5.6356}
    &
    {5.9278}
    &
    {5.7659}
    % &
    % {5.6993}
    % &
    % {5.7659}
\\
    col
    & 
    {3.5597}
    & 
    {3.5511}
    &
    {3.5674}
    &
    {3.5349}
    % &
    % {3.4236}
    % &
    % {3.5349}
\\
    dis
    & 
    {3.9680}
    & 
    {3.9571}
    &
    {4.1632}
    &
    {3.9617}
    % &
    % {3.8632}
    % &
    % {3.9618}
\\
    loud
    & 
    {3.5468}
    & 
    {3.5312}
    &
    {3.8542}
    &
    {3.4789}
    % &
    % {3.4002}
    % &
    % {3.4790}
\\
    pJSD
    & 
    {0.2253}
    & 
    {0.2096}
    &
    {0.1811}
    &
    {0.1770}
    % &
    % {0.2244}
    % &
    % {0.1770}
\\
\bottomrule
\end{tabular}
\label{tab:aux_model}
\end{table}